\documentclass[10pt,twocolumn,letterpaper]{article}
\usepackage[T1]{fontenc}
\usepackage[pagenumbers]{cvpr}
\AtEndPreamble{\crefname{figure}{Fig.}{Figs.}\Crefname{figure}{Fig.}{Figs.}\crefname{table}{Tab.}{Tabs.}\Crefname{table}{Tab.}{Tabs.}}

\usepackage{amsmath,amssymb,booktabs,graphicx,xcolor,xspace,enumitem,multirow,bm,adjustbox}
\xspaceaddexceptions{\% \,}
\usepackage{caption}
\usepackage{placeins}

\usepackage[skip=2pt]{subcaption}
\setlist{leftmargin=1.1em,itemsep=1pt,topsep=2pt,partopsep=0pt,parsep=0pt}
\makeatletter
\renewcommand\paragraph{\@startsection{paragraph}{4}{\z@}{4pt plus 1pt minus 1pt}{-0.6em}{\normalfont\normalsize\bfseries}}
\makeatother
\definecolor{cvprblue}{rgb}{0.21,0.49,0.74}
\usepackage[pagebackref,breaklinks,colorlinks,allcolors=cvprblue]{hyperref}

\def\confName{CVPR}
\def\confYear{2027}

\newcommand{\ours}{Head Anything\xspace}
\newcommand{\pix}{Pixal3D\xspace}
\newcommand{\pixft}{Pixal3D-FT\xspace}
\newcommand{\Phifg}{\Phi_{f\to g}}

\newcommand{\nNgainBestGenCons}{47\xspace}
\newcommand{\nNgainBestGenHead}{32\xspace}

\newcommand{\nRtsFactor}{2.3\xspace}

\newcommand{\nRtsIds}{10\xspace}
\newcommand{\nRtsOurs}{260\xspace}
\newcommand{\nRtsPix}{114\xspace}

\newcommand{\ninvFailOneMm}{0.06\xspace}

\newcommand{\ninvOutGrid}{14.8\xspace}

\newcommand{\ninvResMedTxt}{below 0.01\xspace}

\newcommand{\nnsTargets}{10\xspace}

\title{Head Anything: Correspondence-Guided Joint\\Generation of 3D Heads from Few Images}
\author{Umut Kocasar{\i}$^{1}$ \quad Bart{\l}omiej Baranowski$^{1}$ \quad Richard Shaw$^{2}$\\
Eduardo P\'{e}rez-Pellitero$^{2}$ \quad Matthias Nie{\ss}ner$^{1}$\\
$^{1}$Technical University of Munich, Germany \qquad $^{2}$Huawei Noah's Ark Lab, London, UK}

\begin{document}
\twocolumn[{%
\renewcommand\twocolumn[1][]{#1}%
\maketitle
\vspace*{-10pt}%
{\centering
  \includegraphics[width=\textwidth]{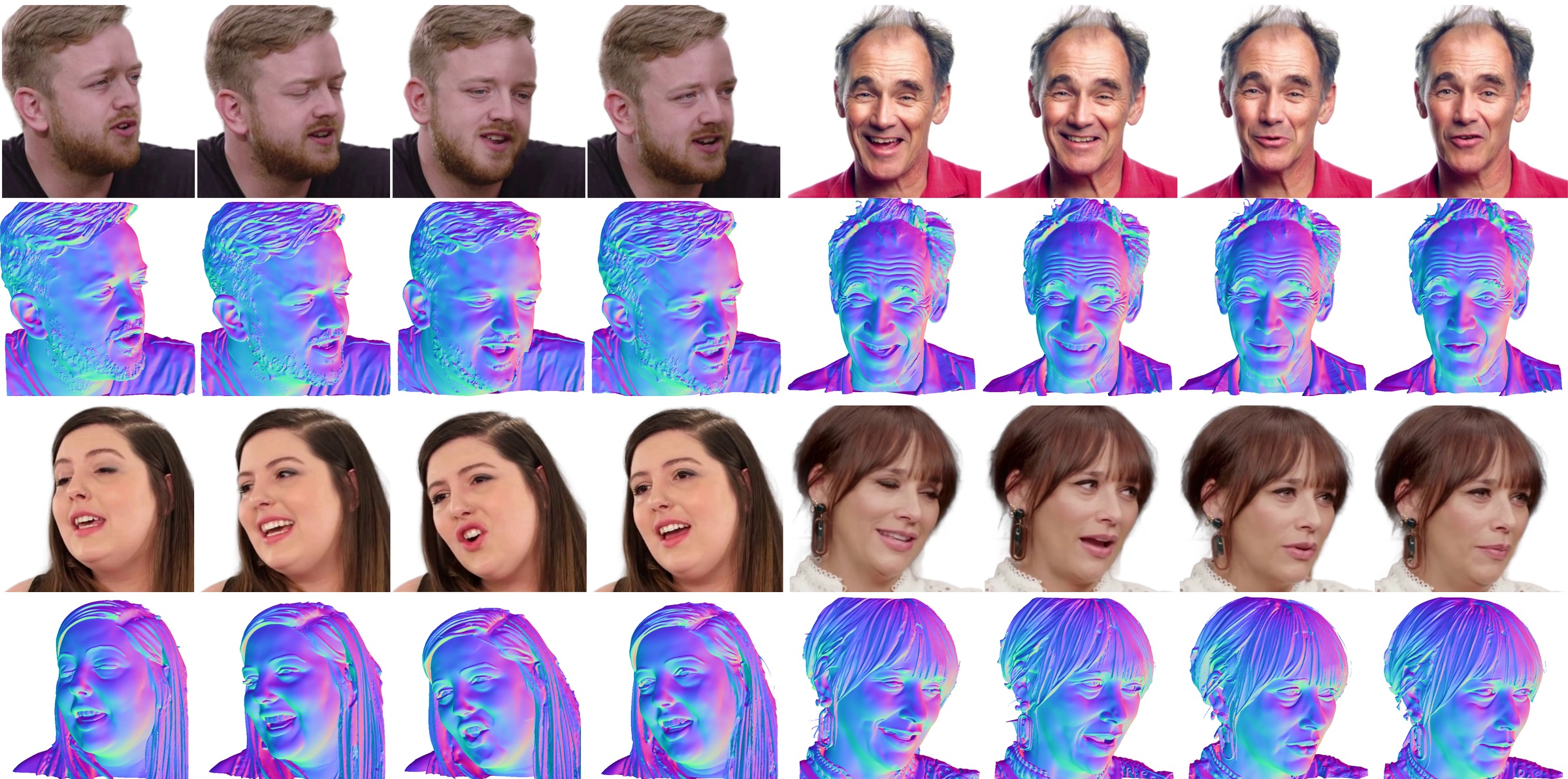}\par
  \captionof{figure}{\textbf{\ours.} From a few in-the-wild video frames of a person, our method jointly generates one detailed mesh per frame, shown below it. Each mesh follows its own image while benefiting from the other views and expressions. Website: \url{https://kocasariumut.github.io/HeadAnything/}}
  \label{fig:teaser}\par}
\vspace{0.9em}
}]
\begin{abstract}
Large image-to-3D generators recover detailed heads from a single photograph but treat every image as a new object.
For several images of one person, they hallucinate unseen regions independently without using information from the other images.
We present \ours, which generates detailed head meshes from a few images with changing expressions and viewpoints.
Instead of training a multi-image prior, we turn a pretrained single-image generator with a native 3D latent space into a joint generator that denoises all images together.
Since global attention between images is hard to learn from scarce 3D head data, lightweight correspondence-attention adapters let every 3D token attend only to its corresponding points in the other images.
We obtain correspondences by fitting a neural parametric head model to the generator's own single-image outputs, requiring no ground truth at test time.
Training on both ground-truth and fitted correspondences makes the adapters robust to fitting errors.
Each mesh stays pixel-aligned with its own image and borrows the unseen regions from all others.
On head scans, \ours lowers the head Chamfer distance of the best image-to-3D generator by \nNgainBestGenHead\% and its cross-frame inconsistency by \nNgainBestGenCons\%, and its meshes improve face tracking.
Code and models will be released.
\end{abstract}

\section{Introduction}
\label{sec:intro}

Telepresence, games, film production and virtual try-on need the 3D geometry of a person's head, including the hair, the ears and the back of the head, at a level of detail that today requires a multi-view capture studio~\cite{kirschstein2023nersemble,martinez2024ava256}.
Most images of people, however, come from casual videos or photo collections, such as those in \cref{fig:teaser}, and depict the head from varying viewpoints and with different expressions.
Recovering complete, detailed and mutually consistent meshes from such images would make studio-quality head geometry available from everyday data, with every mesh following its own image and benefiting from what the others reveal.

This problem is underconstrained.
Each image shows only part of the head, and since expression and pose change, multi-view reconstruction does not apply.
A strong learned prior is indispensable, but high-quality head scans exist for only a few hundred people~\cite{giebenhain2023nphm,wang2022faceverse}, far fewer than the hundreds of thousands of objects used to train general image-to-3D generators~\cite{deitke2023objaverse}.
A prior over multiple images is even harder to learn, since the same point on the head appears at a different position in each image as expressions and poses change, and global attention across images cannot reliably learn these correspondences from so few identities.

Existing methods work around this lack of 3D data with different trade-offs.
Parametric head models~\cite{blanz1999morphable,li2017flame,giebenhain2023nphm} and their trackers~\cite{zielonka2022mica,qian2024vhap,giebenhain2025pixel3dmm,giebenhain2024mononphm} naturally combine many frames, but their geometry is coarse and misses hair and fine detail.
Feed-forward methods predict per-pixel geometry and correspondences~\cite{faceanything}, but only for the visible surface.
Large image-to-3D generators~\cite{xiang2025trellis,xiang2025trellis2,li2025triposg,hunyuan2025v21,wu2025direct3ds2,li2026pixal3d} produce complete, detailed meshes, but from one image at a time, and their multi-image variants assume a static object.
Video and multi-view diffusion models share information along optical flow~\cite{cong2024flatten,geyer2024tokenflow}, epipolar lines~\cite{tseng2023posediffusion,kant2024spad} or a morphable head model~\cite{chen2024morphable,taubner2025cap4d}, but only between 2D images.

\begin{figure*}[t]
  \centering
  \includegraphics[width=\textwidth]{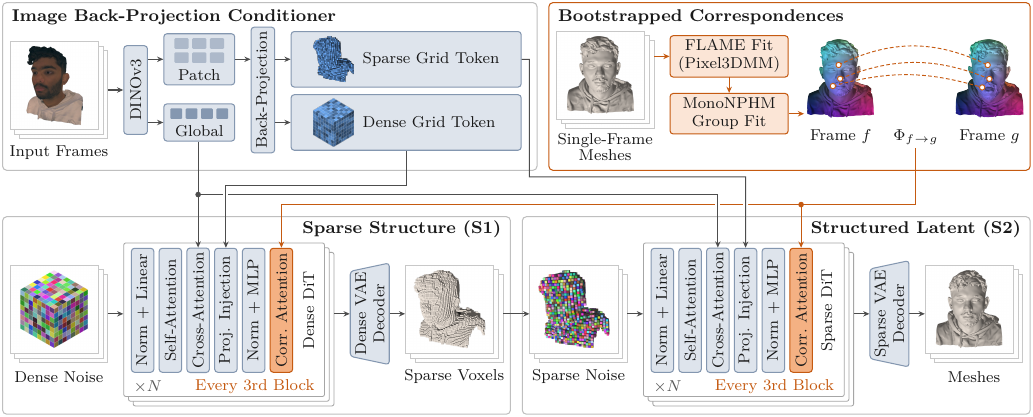}
  \caption{\textbf{Architecture overview.} Each frame conditions the fine-tuned generator \pixft through DINOv3 features back-projected onto the dense (S1) and sparse (S2) grid tokens. All $n$ frames are denoised jointly, and correspondence-attention adapters couple them at the corresponding points $\Phifg$ of a FLAME and MonoNPHM fit to the single-frame meshes. The details of correspondence attention and how it is applied in each stage are shown in \cref{fig:adapter}.}
  \label{fig:pipeline}
\end{figure*}

We introduce \ours for multi-image 3D head generation by adapting a pretrained single-image generator with a native 3D latent space to jointly generate 3D heads from multiple images of the same person.
As \cref{fig:pipeline} shows, it jointly denoises all 3D latents using lightweight adapters letting each 3D token attend only to corresponding points in the other images.
Unlike global attention, correspondence attention knows where to look from the first training step.
The correspondences are deformable and bootstrapped from the generator itself by fitting a neural parametric head model~\cite{giebenhain2023nphm,giebenhain2024mononphm} to its single-image outputs.
This eliminates ground-truth requirements at test time, while training on both ground-truth and fitted correspondences makes the adapters robust to fitting errors.
Routing attention through these correspondences within a 3D latent generator distinguishes our method from image-space correspondence attention and from simply combining a generator with a parametric fit.
Experiments show our meshes are more accurate than all baselines, far more consistent across frames than image-to-3D generators' meshes, and improve a parametric face tracker's performance.
Joint generation yields greater accuracy and consistency than frame-by-frame generation using our backbone, \pix~\cite{li2026pixal3d}, fine-tuned on head scans.

\noindent In summary, our contributions are:
\begin{itemize}
  \item We introduce a joint multi-image head generator that couples the sampling processes of a pretrained single-image 3D generator, so that every mesh follows its own image and benefits from all others.
  \item We propose a novel correspondence attention mechanism in which every 3D token attends to its corresponding points in the other images, given by a neural parametric head model fitted to the generator's own outputs and made robust by mixed ground-truth and fitted correspondences.
  \item We demonstrate that joint generation produces more accurate 3D heads than all baselines, far more consistent unseen regions than image-to-3D generators, and meshes that improve parametric face tracking.
\end{itemize}

\section{Related Work}
\label{sec:related}

\paragraph{Single-image 3D generation.}
3D generators trained on object collections~\cite{deitke2023objaverse} either lift novel-view and multi-view diffusion models to 3D~\cite{liu2023zero123,shi2024mvdream} or generate directly in native 3D latent spaces.
TRELLIS~\cite{xiang2025trellis} denoises a sparse voxel structure and then structured latents on the occupied voxels.
TRELLIS.2~\cite{xiang2025trellis2} makes these latents compact at high resolution.
TripoSG~\cite{li2025triposg} and Hunyuan3D-2.1~\cite{hunyuan2025v21} use rectified flows~\cite{liu2023rectified,lipman2023flow} over vector-set latents, and Direct3D-S2~\cite{wu2025direct3ds2} scales sparse attention to high-resolution volumes.
\pix~\cite{li2026pixal3d} back-projects image features into the voxel grid, so that the generated geometry is aligned with the input pixels and preserves fine details.
All of these methods generate one object from one image, and their multi-image modes assume one rigid object.

\paragraph{3D head reconstruction.}
Morphable models~\cite{blanz1999morphable} such as FLAME~\cite{li2017flame} form the basis of most monocular face reconstruction and tracking methods~\cite{feng2021deca,zielonka2022mica,qian2024vhap,giebenhain2025pixel3dmm}.
They provide a consistent topology, but their geometry is coarse and covers neither hair nor accessories.
NPHM~\cite{giebenhain2023nphm} models identity and expression of complete heads with neural fields learned from scans, and MonoNPHM~\cite{giebenhain2024mononphm} adds appearance and fits it to monocular videos, with detail bounded by its latent space.
Feed-forward methods regress per-frame geometry and correspondences from images~\cite{faceanything}, but only for the visible surface.
Capture setups~\cite{kirschstein2023nersemble,martinez2024ava256} and the avatars built on them~\cite{qian2024gaussianavatars} need calibrated camera rigs.
We use a parametric head model only as a source of dense correspondences, while the geometry comes from the generator.

\paragraph{Correspondence-guided multi-image generation.}
Video diffusion models attend between frames at the same spatial location~\cite{blattmann2023alignlatents,wu2023tuneavideo}, while video editing methods follow correspondences.
FLATTEN restricts attention to patches on the same optical-flow path~\cite{cong2024flatten}, and TokenFlow propagates diffusion features through nearest-neighbour correspondences~\cite{geyer2024tokenflow}.
Multi-view diffusion models attend along epipolar lines~\cite{tseng2023posediffusion,kant2024spad}, and for heads, Morphable Diffusion~\cite{chen2024morphable} and CAP4D~\cite{taubner2025cap4d} condition multi-view diffusion on a morphable model and reconstruct an avatar from the generated views.
4D methods reconstruct dynamic objects from video~\cite{jiang2024consistent4d,xie2025sv4d,ren2024l4gm} or animate one generated mesh~\cite{chen2025v2m4}, and training-free approaches average several diffusion processes where they overlap~\cite{bartal2023multidiffusion,kim2024synctweedies}.
All correspondence-guided methods above exchange information between 2D images, with correspondences from optical flow, camera geometry or a rendered morphable model, and obtain 3D only through a separate reconstruction.
We instead exchange information between sparse tokens within a native 3D latent generator.
Deformable correspondences bootstrapped from the generator's single-image outputs connect tokens across different head poses and expressions.
Each frame retains its own detailed mesh.
As in ControlNet~\cite{zhang2023controlnet}, the adapters are zero-initialised modules on a pretrained backbone, which keeps its prior intact at the start of training.

\section{Method}
\label{sec:method}
\begin{figure*}[t]
  \centering
  \includegraphics[width=\textwidth]{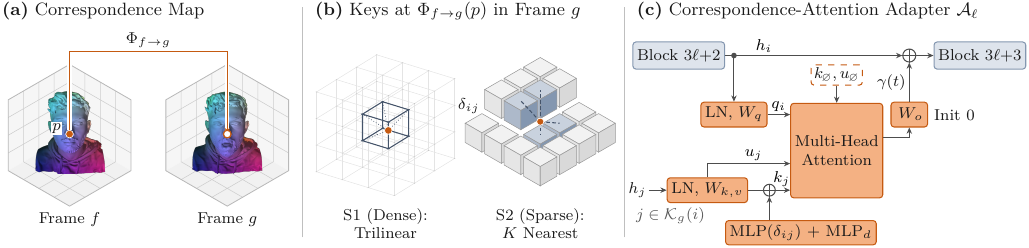}
  \caption{\textbf{Correspondence attention.} (a)~$\Phifg$ maps a token position $p$ of frame $f$ to its corresponding point in frame $g$. (b)~Keys are sampled trilinearly in S1 and taken from the $K$ nearest occupied tokens in S2. (c)~The adapter attends to these keys alongside a learned null key-value pair, then adds its gated output to the residual stream.}
  \label{fig:adapter}
\end{figure*}

Given $n$ images $\{I_f\}_{f=1}^{n}$ of a person's head, we generate one detailed mesh $M_f$ per image, pixel-aligned with $I_f$ and completed from $I_{g\neq f}$ in regions unseen in $I_f$.
We fine-tune a single-image generator on head scans (\cref{sec:prelim,sec:finetune}) and couple $n$ of its sampling processes (\cref{sec:joint}) through correspondence attention (\cref{sec:corrattn}), routed by correspondences bootstrapped from its own outputs (\cref{sec:fitting}).

\subsection{Preliminaries}
\label{sec:prelim}
We build on \pix~\cite{li2026pixal3d}, which generates a mesh in two stages.
A \emph{sparse-structure} flow (S1) denoises a dense $16^3{\times}8$ latent, which a dense VAE decoder maps to occupied voxels on a $64^3$ grid.
A \emph{structured-latent} flow (S2) then denoises latents at these voxels at $512^3$ resolution, followed by cascaded refinement to $1024^3$ and mesh extraction with a sparse VAE decoder.
Both flows use diffusion transformers~\cite{peebles2023dit} with $N{=}30$ blocks.
They receive \mbox{DINOv3}~\cite{simeoni2025dinov3} features through cross-attention and projection injection, which back-projects patch features onto grid tokens to align the output with the input pixels.
Each flow is trained with the rectified-flow objective~\cite{liu2023rectified,lipman2023flow}
\begin{equation}
  \mathcal{L}(\theta) = \mathbb{E}_{t,\,x_0,\,\epsilon}\, \big\| v_\theta(x_t, t, c) - (\epsilon - x_0) \big\|^2 ,
  \label{eq:fm}
\end{equation}
where $x_t = (1-t)\,x_0 + t\,\epsilon$ interpolates a clean latent $x_0$ and Gaussian noise $\epsilon$ at flow time $t\in[0,1]$, $c$ is the image condition and $v_\theta$ the predicted velocity.

\subsection{A Single-Image Head Prior}
\label{sec:finetune}
Object-level generators miss the specific shape of a face and often fail on hair, so we fine-tune \pix on photorealistic renders of the NPHM head scans~\cite{giebenhain2023nphm}.
The resulting \pixft is a strong single-image head prior, but it generates every frame on its own, so the frames do not benefit from each other. It also hallucinates the back of the head and the hair independently for every frame.

\subsection{Joint Generation}
\label{sec:joint}
Instead of $n$ independent samples, our method draws the latents of all frames as one sample from their joint distribution given all images $\{I_f\}_{f=1}^{n}$.
Each frame keeps its own noise, condition $c_f$, tokens and, in S2, occupied voxels, so that every mesh stays aligned with its own image.
The \pix sampler advances all frames in lockstep.
The \mbox{transformer} blocks process each frame independently.
After block $3\ell{+}2$ in both flows, with $\ell=0,\dots,9$, an adapter $\mathcal{A}_\ell$ updates the tokens $H_f$ of each frame $f$ using the tokens $H_g$ of the other frames $g\neq f$,
\begin{equation}
  H_f \leftarrow H_f + \gamma(t)\, W_o\, \mathrm{Attn}_\ell\big(H_f, \{H_g\}_{g\neq f}, \Phi\big) ,
  \label{eq:residual}
\end{equation}
where $\mathrm{Attn}_\ell$ is the correspondence attention of the adapter, $\Phi$ the correspondences between the frames, $W_o$ a zero-initialised output projection and $\gamma(t)$ a learned gate on the timestep embedding.
At initialisation, the joint generator thus equals the single-image generator. Since both stages are joint, the occupied voxels of S1 already benefit from the other frames.
The backbone is fine-tuned together with the adapters at a lower learning rate, so that it can adapt to the incoming information without forgetting its prior.

\begin{figure*}[t]
  \centering
  {\setlength{\tabcolsep}{1.00pt}\renewcommand{\arraystretch}{0}
\newlength{\qcwreal}\setlength{\qcwreal}{\dimexpr(\linewidth-20.0pt)/11\relax}
\begin{tabular}{@{}ccccccccccc@{}}
\parbox[b][2.15em][c]{\qcwreal}{\centering\fontsize{6.2pt}{7.2pt}\selectfont \adjustbox{max width=\qcwreal}{Input}} & \parbox[b][2.15em][c]{\qcwreal}{\centering\fontsize{6.2pt}{7.2pt}\selectfont \adjustbox{max width=\qcwreal}{MonoNPHM}\\[-0.5pt]\cite{giebenhain2024mononphm}} & \parbox[b][2.15em][c]{\qcwreal}{\centering\fontsize{6.2pt}{7.2pt}\selectfont \adjustbox{max width=\qcwreal}{Pixel3DMM}\\[-0.5pt]\cite{giebenhain2025pixel3dmm}} & \parbox[b][2.15em][c]{\qcwreal}{\centering\fontsize{6.2pt}{7.2pt}\selectfont \adjustbox{max width=\qcwreal}{VHAP}\\[-0.5pt]\cite{qian2024vhap}} & \parbox[b][2.15em][c]{\qcwreal}{\centering\fontsize{6.2pt}{7.2pt}\selectfont \adjustbox{max width=\qcwreal}{TRELLIS}\\[-0.5pt]\cite{xiang2025trellis}} & \parbox[b][2.15em][c]{\qcwreal}{\centering\fontsize{6.2pt}{7.2pt}\selectfont \adjustbox{max width=\qcwreal}{Direct3D-S2}\\[-0.5pt]\cite{wu2025direct3ds2}} & \parbox[b][2.15em][c]{\qcwreal}{\centering\fontsize{6.2pt}{7.2pt}\selectfont \adjustbox{max width=\qcwreal}{TRELLIS.2}\\[-0.5pt]\cite{xiang2025trellis2}} & \parbox[b][2.15em][c]{\qcwreal}{\centering\fontsize{6.2pt}{7.2pt}\selectfont \adjustbox{max width=\qcwreal}{TripoSG}\\[-0.5pt]\cite{li2025triposg}} & \parbox[b][2.15em][c]{\qcwreal}{\centering\fontsize{6.2pt}{7.2pt}\selectfont \adjustbox{max width=\qcwreal}{Hunyuan3D-2.1}\\[-0.5pt]\cite{hunyuan2025v21}} & \parbox[b][2.15em][c]{\qcwreal}{\centering\fontsize{6.2pt}{7.2pt}\selectfont \adjustbox{max width=\qcwreal}{Pixal3D}\\[-0.5pt]\cite{li2026pixal3d}} & \parbox[b][2.15em][c]{\qcwreal}{\centering\fontsize{6.2pt}{7.2pt}\selectfont \adjustbox{max width=\qcwreal}{Ours}} \\[1.5pt]
\includegraphics[width=\qcwreal]{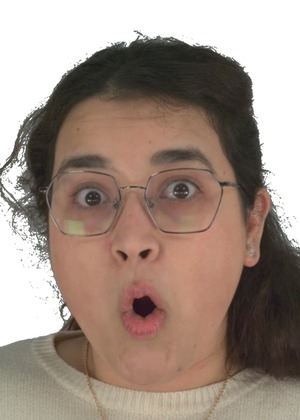} & \includegraphics[width=\qcwreal]{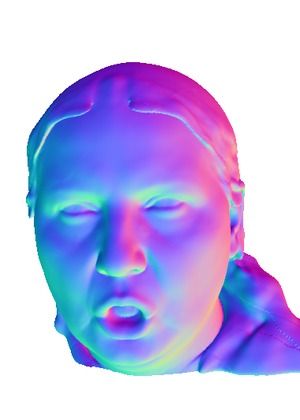} & \includegraphics[width=\qcwreal]{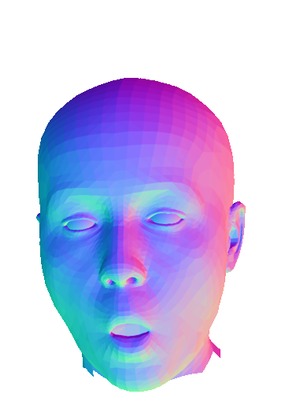} & \includegraphics[width=\qcwreal]{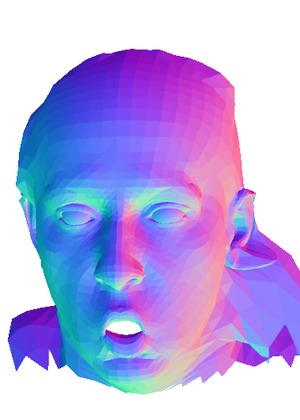} & \includegraphics[width=\qcwreal]{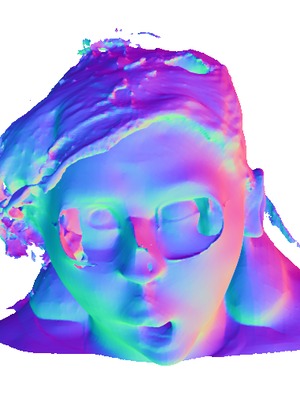} & \includegraphics[width=\qcwreal]{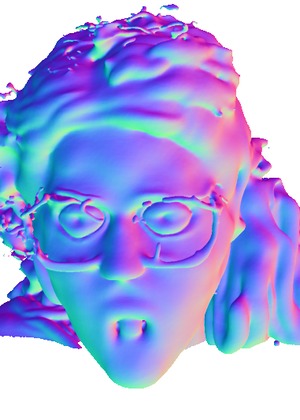} & \includegraphics[width=\qcwreal]{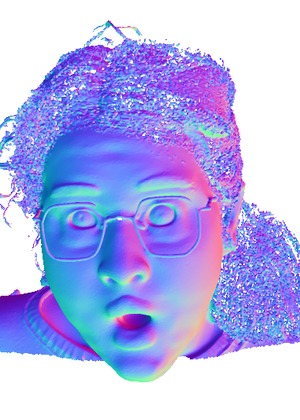} & \includegraphics[width=\qcwreal]{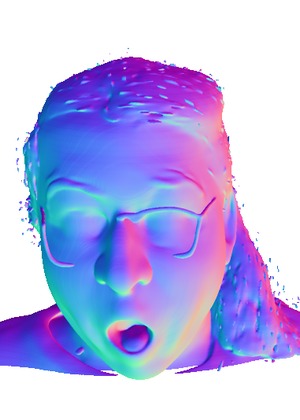} & \includegraphics[width=\qcwreal]{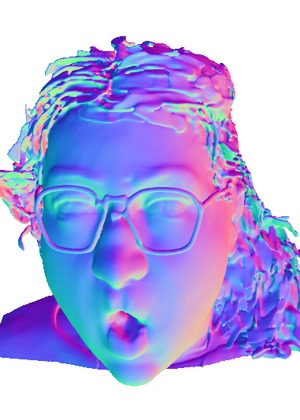} & \includegraphics[width=\qcwreal]{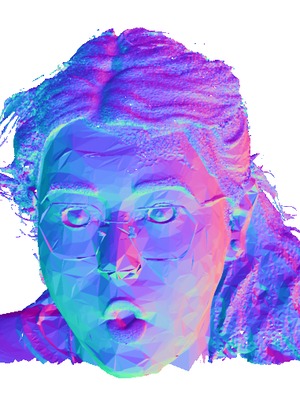} & \includegraphics[width=\qcwreal]{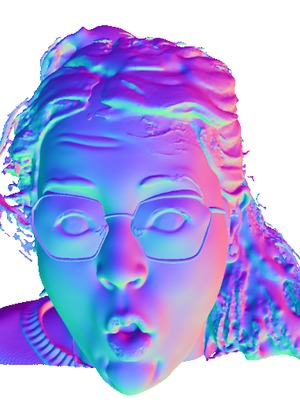} \\[2.0pt]
\includegraphics[width=\qcwreal]{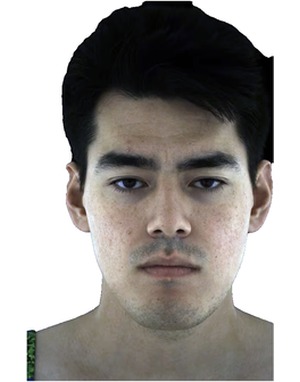} & \includegraphics[width=\qcwreal]{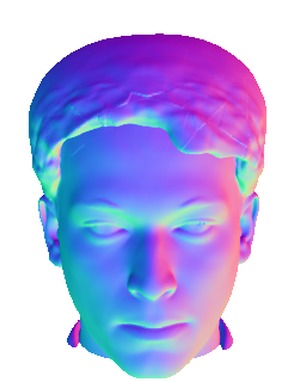} & \includegraphics[width=\qcwreal]{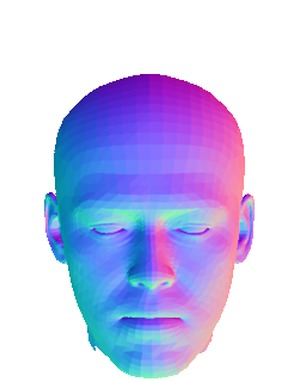} & \includegraphics[width=\qcwreal]{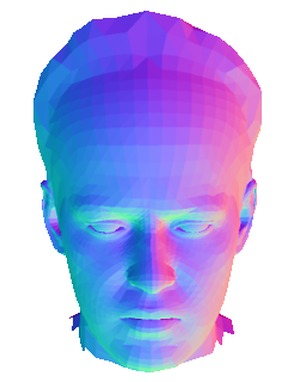} & \includegraphics[width=\qcwreal]{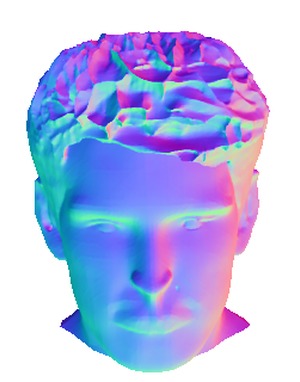} & \includegraphics[width=\qcwreal]{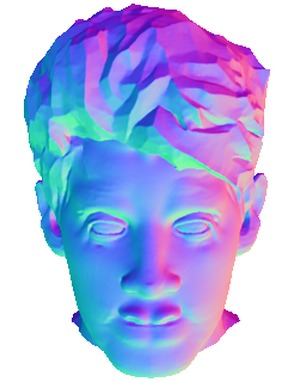} & \includegraphics[width=\qcwreal]{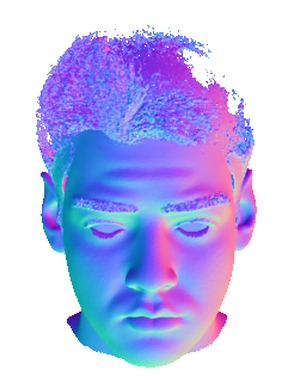} & \includegraphics[width=\qcwreal]{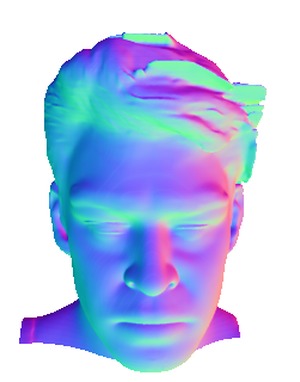} & \includegraphics[width=\qcwreal]{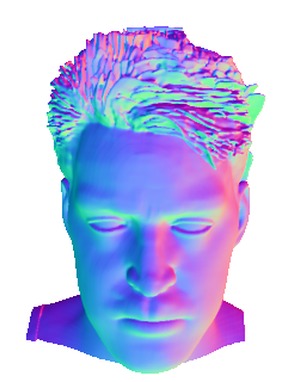} & \includegraphics[width=\qcwreal]{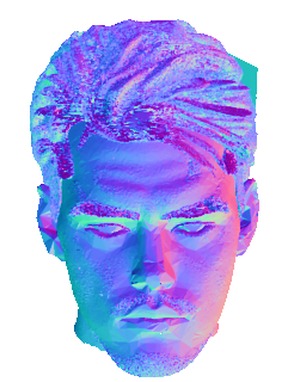} & \includegraphics[width=\qcwreal]{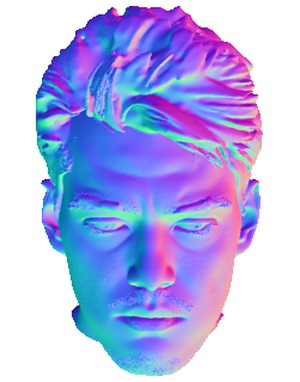} \\[2.0pt]
\includegraphics[width=\qcwreal]{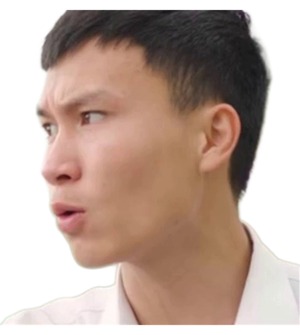} & \includegraphics[width=\qcwreal]{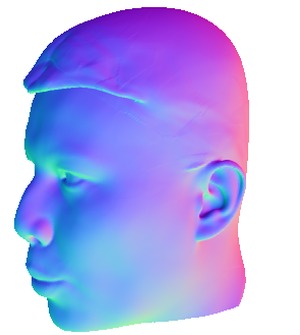} & \includegraphics[width=\qcwreal]{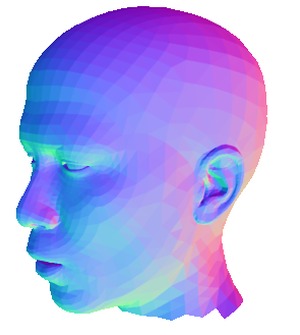} & \includegraphics[width=\qcwreal]{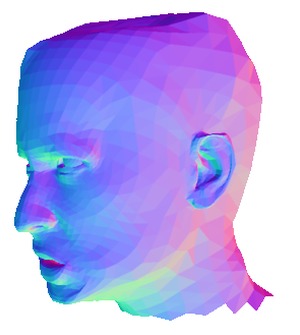} & \includegraphics[width=\qcwreal]{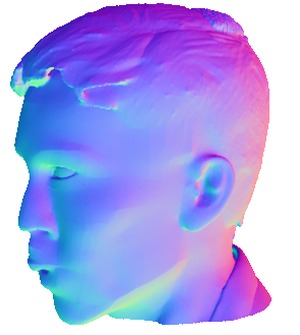} & \includegraphics[width=\qcwreal]{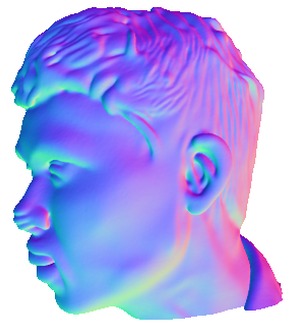} & \includegraphics[width=\qcwreal]{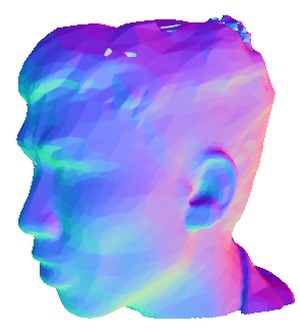} & \includegraphics[width=\qcwreal]{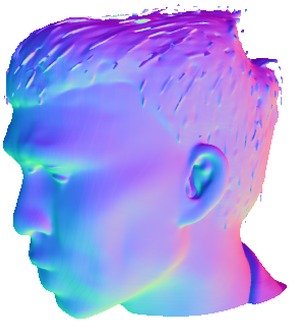} & \includegraphics[width=\qcwreal]{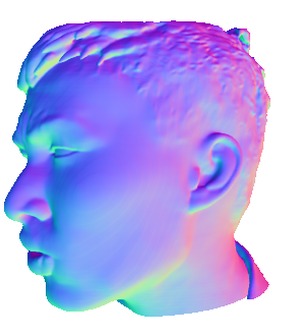} & \includegraphics[width=\qcwreal]{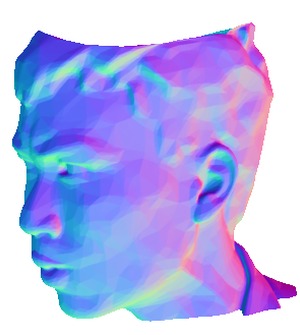} & \includegraphics[width=\qcwreal]{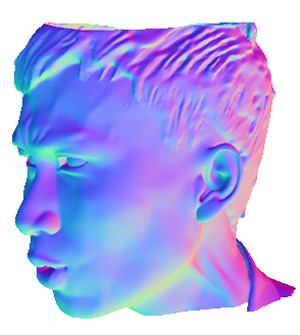} \\[2.0pt]
\includegraphics[width=\qcwreal]{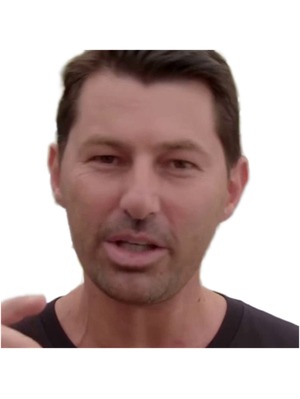} & \includegraphics[width=\qcwreal]{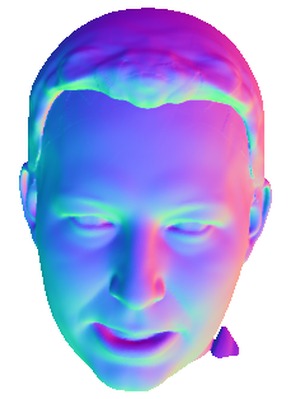} & \includegraphics[width=\qcwreal]{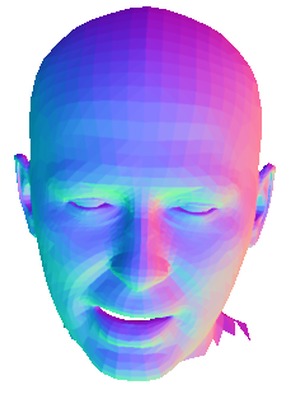} & \includegraphics[width=\qcwreal]{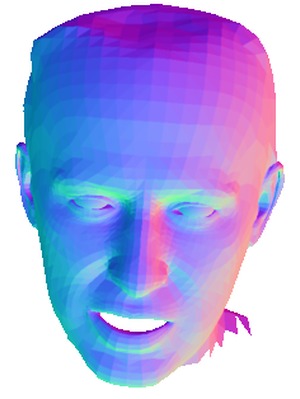} & \includegraphics[width=\qcwreal]{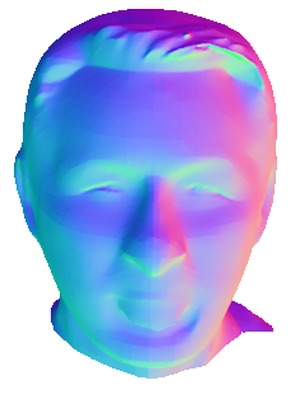} & \includegraphics[width=\qcwreal]{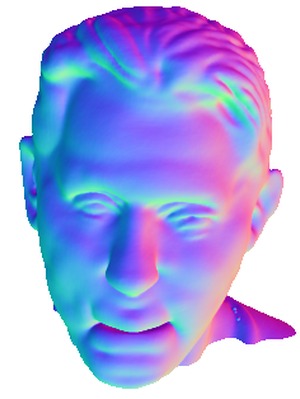} & \includegraphics[width=\qcwreal]{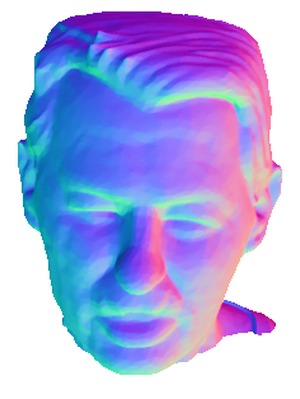} & \includegraphics[width=\qcwreal]{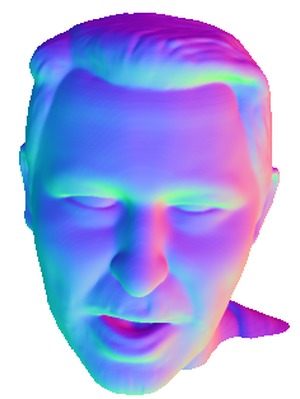} & \includegraphics[width=\qcwreal]{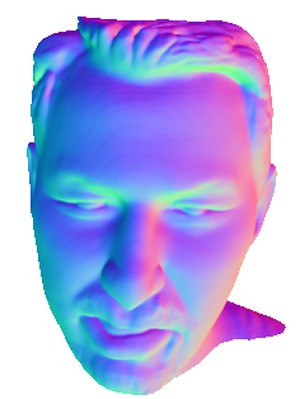} & \includegraphics[width=\qcwreal]{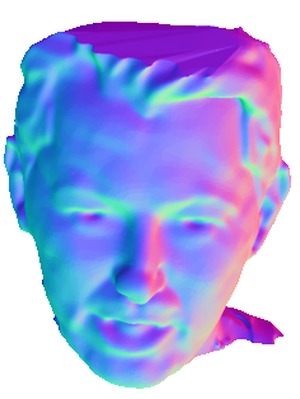} & \includegraphics[width=\qcwreal]{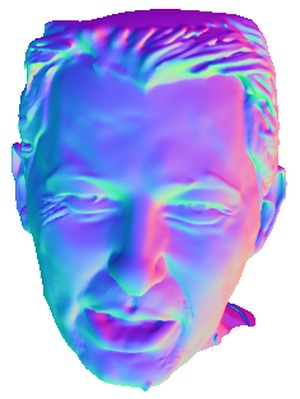} \\
\end{tabular}
}
  \caption{\textbf{Mesh generation comparison on NeRSemble, AVA-256, VFHQ and CelebV-HQ.} Our meshes, rendered with surface normals in the input camera, have a better surface quality than those of the baselines.}
  \label{fig:real}
\end{figure*}

\subsection{Correspondence Attention}
\label{sec:corrattn}
Different frames observe different regions of the head, so the tokens of a region that one image does not show can take their shape from the frames that do.
Since the head turns and the face deforms between frames, the same voxel holds different points of the head in different frames, and a token can only profit from another frame if it knows where its corresponding point lies.
Global attention to all tokens of other frames would have to learn from a few hundred identities where to attend among tens of thousands of tokens, which it cannot do reliably (\cref{sec:ablation}).
We thus restrict attention to corresponding positions, as \cref{fig:adapter} illustrates, so the adapters only learn what to take from each frame.

\paragraph{Correspondence map.}
Let $A_f$ map points from the parametric head model's coordinate system into the latent grid of image frame $f$.
Let $\varphi_{f\to g}$ map a head point in frame $f$ to its corresponding point in frame $g$, with both points expressed in head-model coordinates.
A grid point $p$ of frame $f$ is mapped into frame $g$ by
\begin{equation}
  \Phifg(p) = A_g\, \varphi_{f\to g}\big(A_f^{-1} p\big) .
  \label{eq:phidef}
\end{equation}
For training groups with ground-truth correspondences, $A_f$ is known from the scan and $\varphi_{f\to g}$ comes from the scan registrations $r_f\in\mathbb{R}^{V\times 3}$, whose vertices with the same index $v$ are corresponding points,
\begin{equation}
  \varphi_{f\to g}(y) = y + \textstyle\sum_{v\in\mathcal{N}(y)} w_v\,(r_{g,v} - r_{f,v}) ,
  \label{eq:phi}
\end{equation}
where $\mathcal{N}(y)$ are the nearest vertices of $y$ and $w_v$ normalised inverse-distance weights, so points off the surface, such as hair and clothing, follow the nearest surface and the map covers the whole head.
At test time, $A_f$ and $\varphi_{f\to g}$ come from a parametric model fitted to the generator's outputs (\cref{sec:fitting}).
We use NPHM~\cite{giebenhain2023nphm} rather than FLAME~\cite{li2017flame}, since it gives accurate correspondences on the complete head, including the scalp, hair and neck that the other frames have to complete.

\paragraph{Keys at corresponding positions.}
Token $i$ of frame $f$ at position $p_i$ forms the query $q_i = W_q\,\mathrm{LN}(h_i)$, with layer normalisation $\mathrm{LN}$, and gathers a key set $\mathcal{K}_g(i)$ at $\Phifg(p_i)$ in every other frame $g$.
The S1 latent is a dense $16^3$ grid, so its keys and values are sampled trilinearly at $\Phifg(p_i)$, with an embedding of whether the correspondence is valid and the depth features defined below.
The S2 tokens are sparse, so we take the $K{=}4$ nearest occupied tokens $j$ of frame $g$ at $p_j$, with values $u_j = W_v\,\mathrm{LN}(h_j)$ and keys that embed the offset $\delta_{ij} = p_j - \Phifg(p_i)$ and two depths,
\begin{equation}
  k_j = W_k\,\mathrm{LN}(h_j) + \mathrm{MLP}(\delta_{ij}) + \mathrm{MLP}_{d}(d_i, d_j) ,
  \label{eq:keys}
\end{equation}
where $d_i$ and $d_j$ are the coordinates of the query and the key along the viewing axes of their frames' latent grids.
These depth features indicate on which side of the head a point lies as seen from each frame, so the adapter can learn which frames to rely on.
A learned null key-value pair $(k_\varnothing, u_\varnothing)$ completes the set, allowing tokens to ignore occluded or incorrect correspondences; those outside the grid or without a valid fit are masked.
Multi-head attention over these $(n{-}1)K{+}1$ keys gives $\mathrm{Attn}_\ell$, which connects corresponding points from the first training step.
With $T$ tokens per frame and $K \ll T$, a token attends to $(n{-}1)K{+}1$ keys instead of all $(n{-}1)T$ tokens, so the total cost of $\mathcal{O}(n(n{-}1)KT)$ is a factor $T/K$ below global attention.

\begin{figure*}[t]
  \centering
  {\setlength{\tabcolsep}{1.00pt}\renewcommand{\arraystretch}{0}
\newlength{\qcwbench}\setlength{\qcwbench}{\dimexpr(\linewidth-22.0pt)/12\relax}
\begin{tabular}{@{}cccccccccccc@{}}
\parbox[b][2.15em][c]{\qcwbench}{\centering\fontsize{6.2pt}{7.2pt}\selectfont \adjustbox{max width=\qcwbench}{Input}} & \parbox[b][2.15em][c]{\qcwbench}{\centering\fontsize{6.2pt}{7.2pt}\selectfont \adjustbox{max width=\qcwbench}{MonoNPHM}\\[-0.5pt]\cite{giebenhain2024mononphm}} & \parbox[b][2.15em][c]{\qcwbench}{\centering\fontsize{6.2pt}{7.2pt}\selectfont \adjustbox{max width=\qcwbench}{Pixel3DMM}\\[-0.5pt]\cite{giebenhain2025pixel3dmm}} & \parbox[b][2.15em][c]{\qcwbench}{\centering\fontsize{6.2pt}{7.2pt}\selectfont \adjustbox{max width=\qcwbench}{VHAP}\\[-0.5pt]\cite{qian2024vhap}} & \parbox[b][2.15em][c]{\qcwbench}{\centering\fontsize{6.2pt}{7.2pt}\selectfont \adjustbox{max width=\qcwbench}{TRELLIS}\\[-0.5pt]\cite{xiang2025trellis}} & \parbox[b][2.15em][c]{\qcwbench}{\centering\fontsize{6.2pt}{7.2pt}\selectfont \adjustbox{max width=\qcwbench}{Direct3D-S2}\\[-0.5pt]\cite{wu2025direct3ds2}} & \parbox[b][2.15em][c]{\qcwbench}{\centering\fontsize{6.2pt}{7.2pt}\selectfont \adjustbox{max width=\qcwbench}{TRELLIS.2}\\[-0.5pt]\cite{xiang2025trellis2}} & \parbox[b][2.15em][c]{\qcwbench}{\centering\fontsize{6.2pt}{7.2pt}\selectfont \adjustbox{max width=\qcwbench}{TripoSG}\\[-0.5pt]\cite{li2025triposg}} & \parbox[b][2.15em][c]{\qcwbench}{\centering\fontsize{6.2pt}{7.2pt}\selectfont \adjustbox{max width=\qcwbench}{Hunyuan3D-2.1}\\[-0.5pt]\cite{hunyuan2025v21}} & \parbox[b][2.15em][c]{\qcwbench}{\centering\fontsize{6.2pt}{7.2pt}\selectfont \adjustbox{max width=\qcwbench}{Pixal3D}\\[-0.5pt]\cite{li2026pixal3d}} & \parbox[b][2.15em][c]{\qcwbench}{\centering\fontsize{6.2pt}{7.2pt}\selectfont \adjustbox{max width=\qcwbench}{Ours}} & \parbox[b][2.15em][c]{\qcwbench}{\centering\fontsize{6.2pt}{7.2pt}\selectfont \adjustbox{max width=\qcwbench}{GT scan}} \\[1.5pt]
\includegraphics[width=\qcwbench]{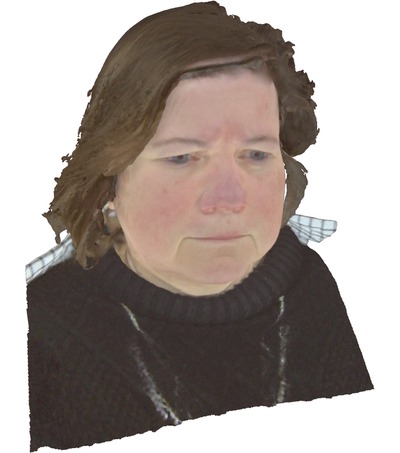} & \includegraphics[width=\qcwbench]{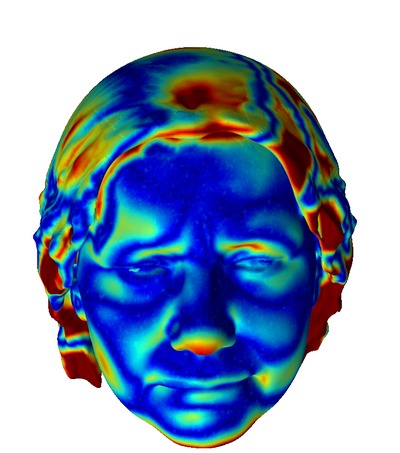} & \includegraphics[width=\qcwbench]{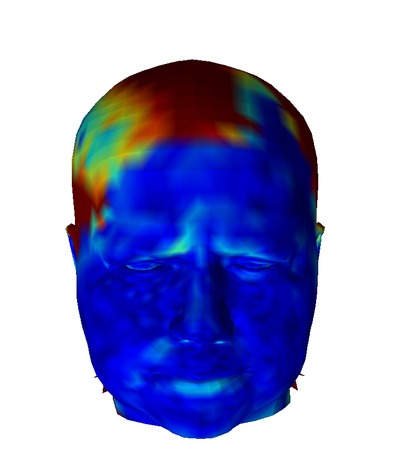} & \includegraphics[width=\qcwbench]{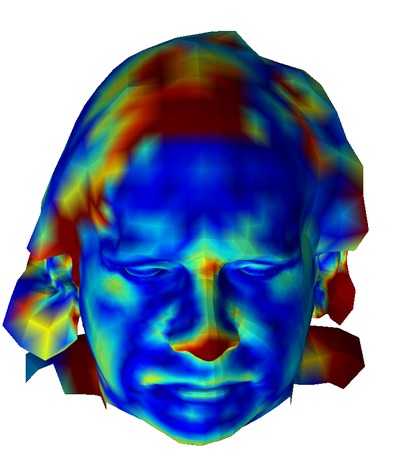} & \includegraphics[width=\qcwbench]{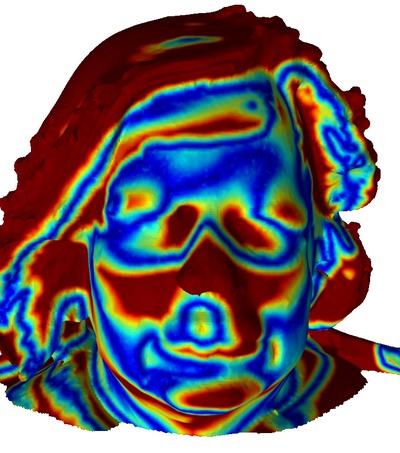} & \includegraphics[width=\qcwbench]{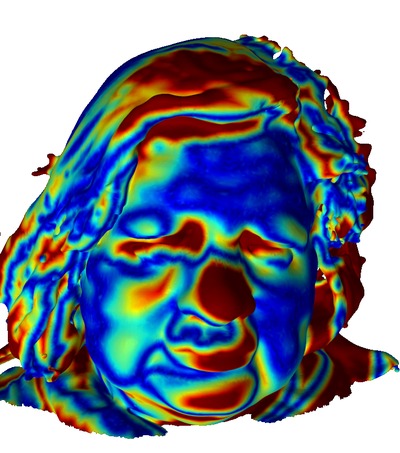} & \includegraphics[width=\qcwbench]{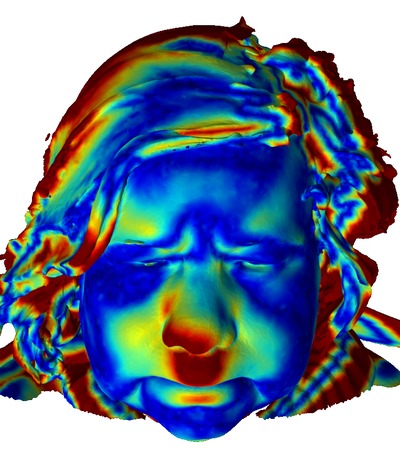} & \includegraphics[width=\qcwbench]{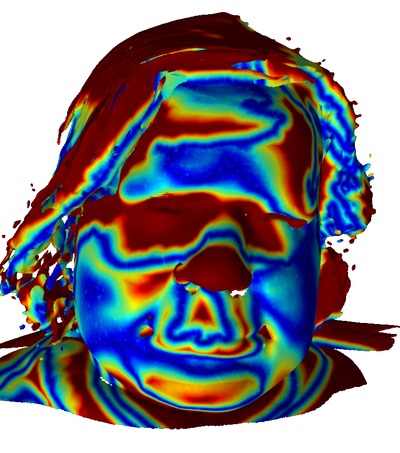} & \includegraphics[width=\qcwbench]{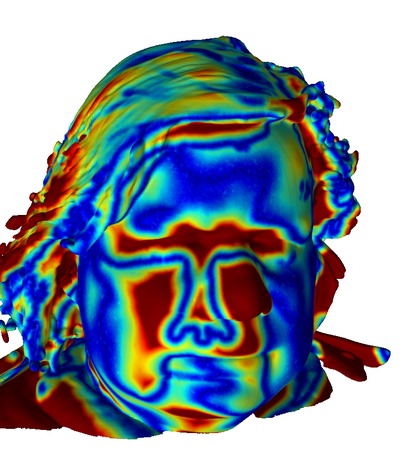} & \includegraphics[width=\qcwbench]{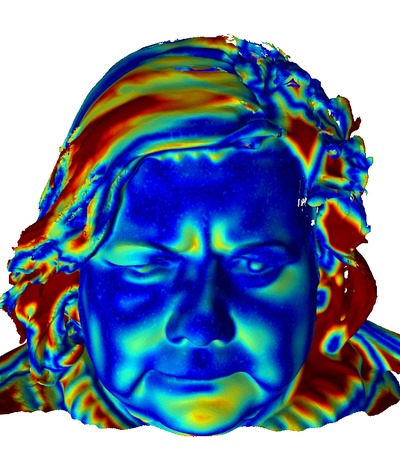} & \includegraphics[width=\qcwbench]{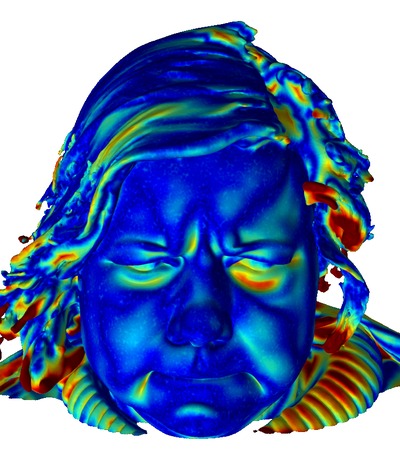} & \includegraphics[width=\qcwbench]{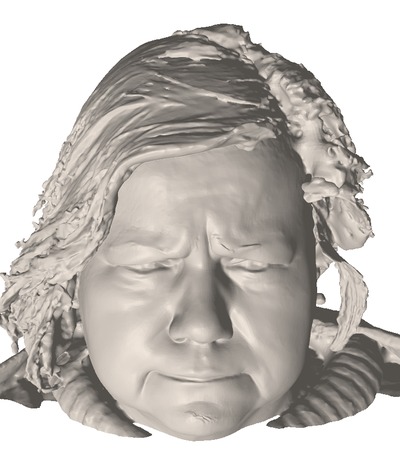} \\[2.0pt]
\includegraphics[width=\qcwbench]{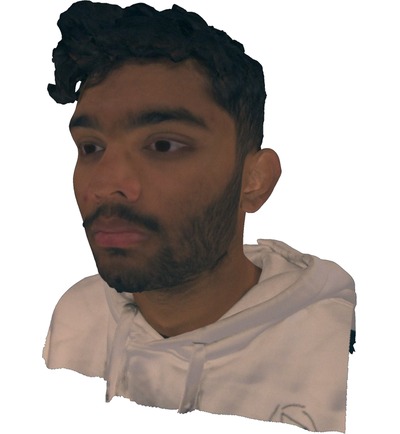} & \includegraphics[width=\qcwbench]{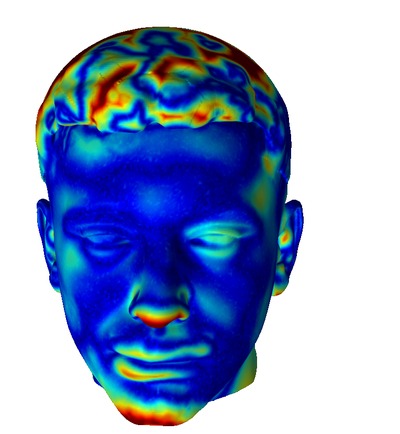} & \includegraphics[width=\qcwbench]{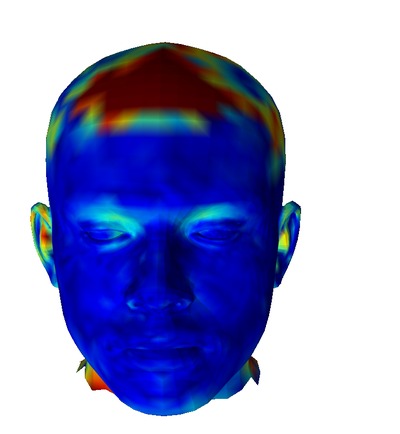} & \includegraphics[width=\qcwbench]{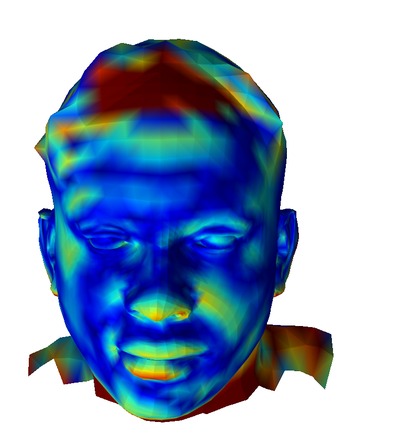} & \includegraphics[width=\qcwbench]{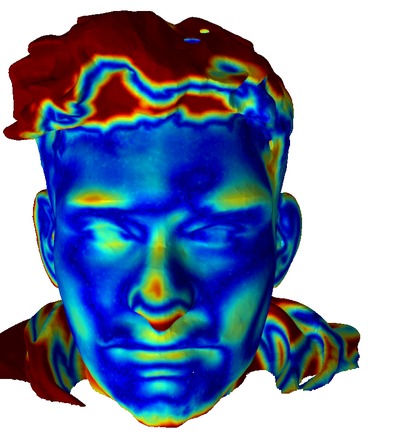} & \includegraphics[width=\qcwbench]{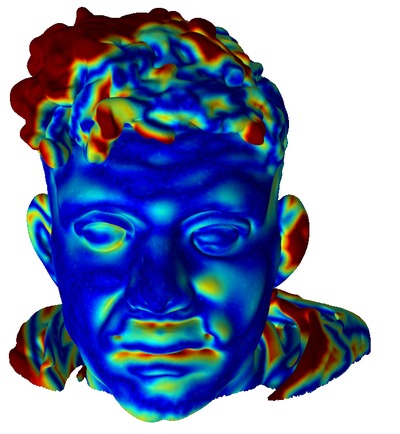} & \includegraphics[width=\qcwbench]{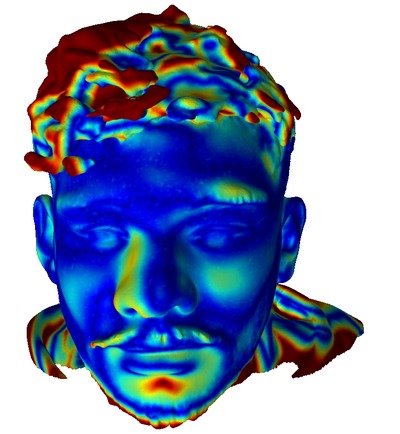} & \includegraphics[width=\qcwbench]{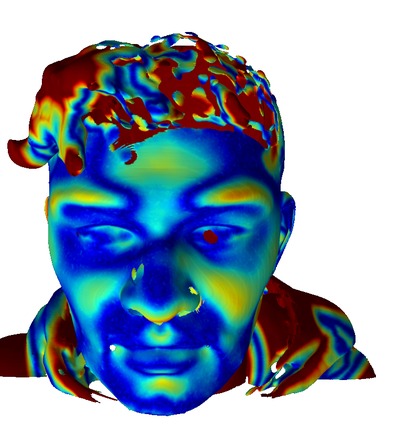} & \includegraphics[width=\qcwbench]{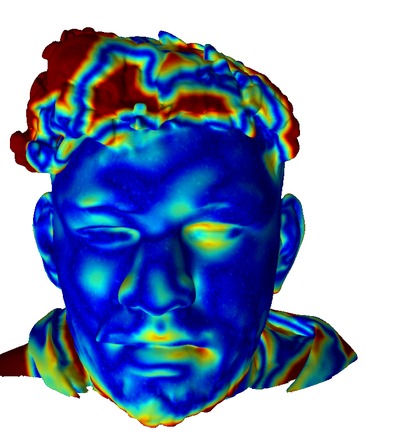} & \includegraphics[width=\qcwbench]{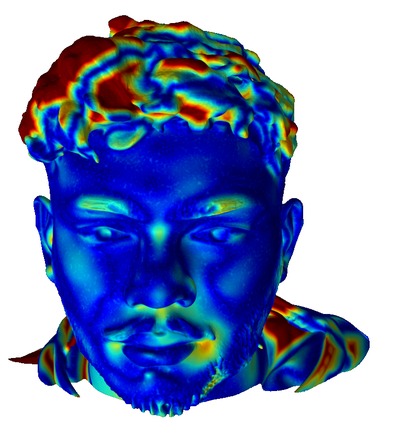} & \includegraphics[width=\qcwbench]{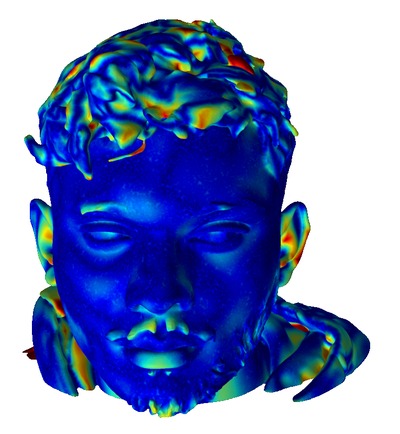} & \includegraphics[width=\qcwbench]{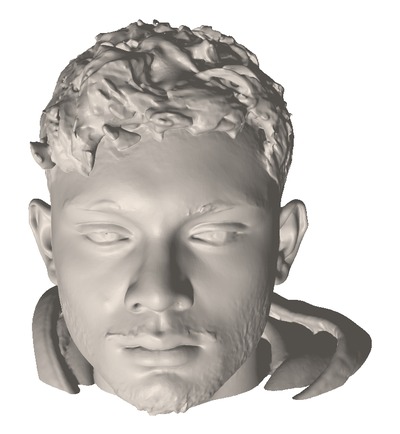} \\[2.0pt]
\includegraphics[width=\qcwbench]{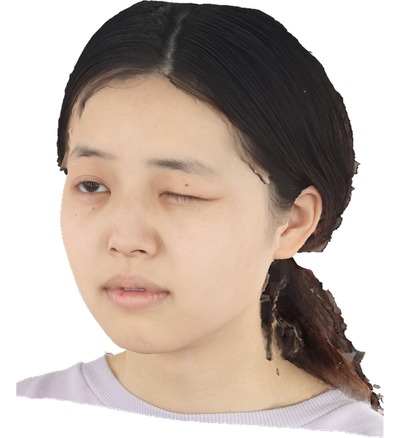} & \includegraphics[width=\qcwbench]{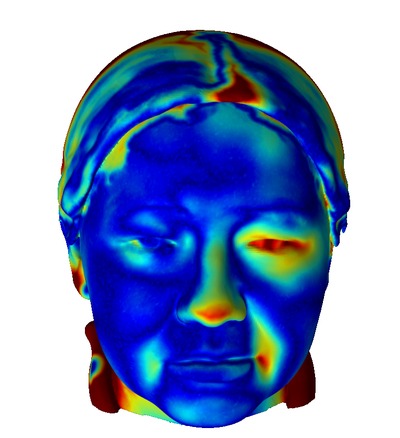} & \includegraphics[width=\qcwbench]{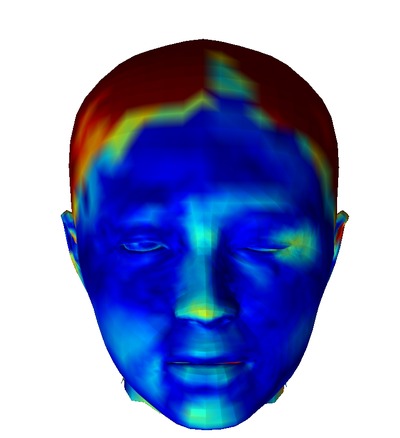} & \includegraphics[width=\qcwbench]{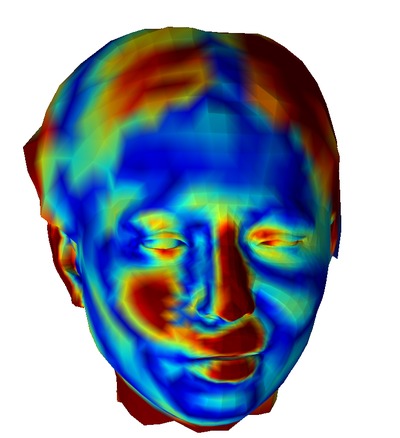} & \includegraphics[width=\qcwbench]{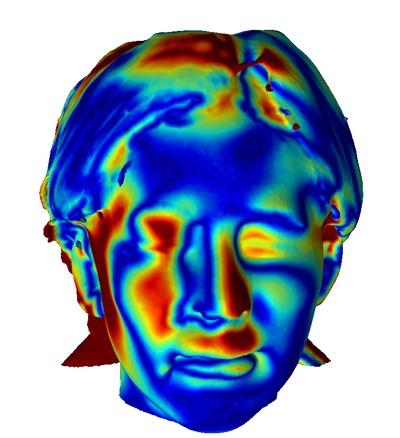} & \includegraphics[width=\qcwbench]{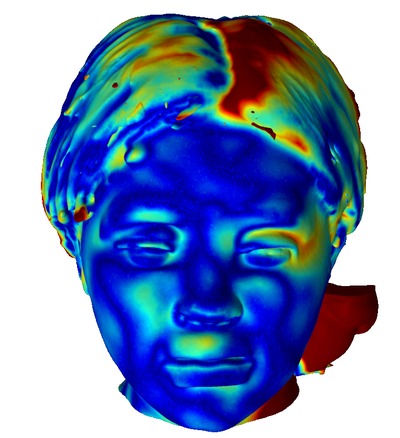} & \includegraphics[width=\qcwbench]{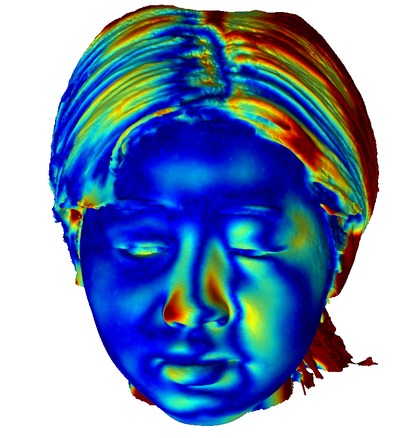} & \includegraphics[width=\qcwbench]{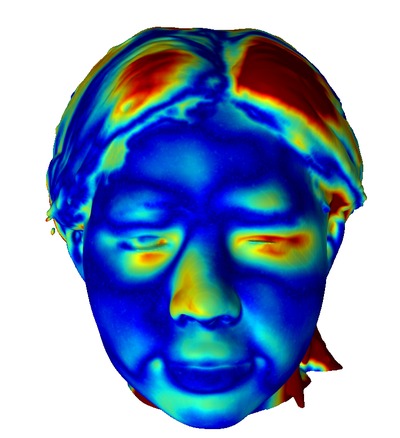} & \includegraphics[width=\qcwbench]{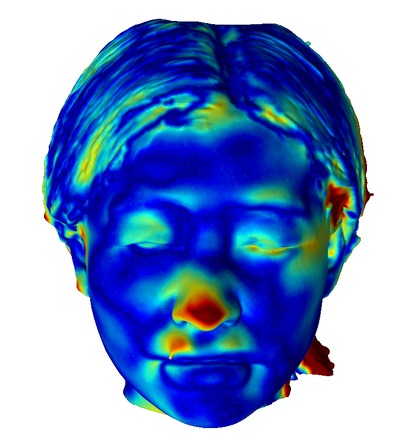} & \includegraphics[width=\qcwbench]{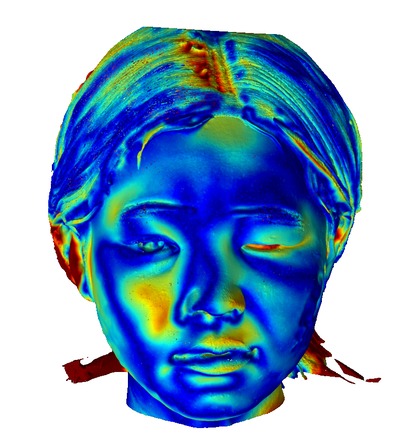} & \includegraphics[width=\qcwbench]{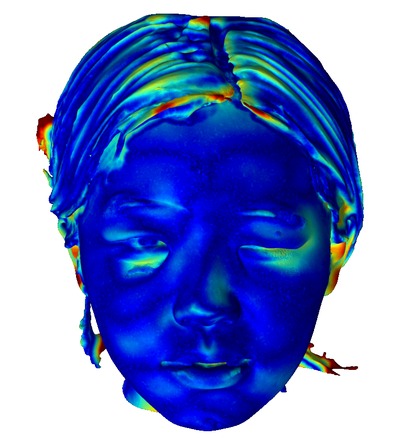} & \includegraphics[width=\qcwbench]{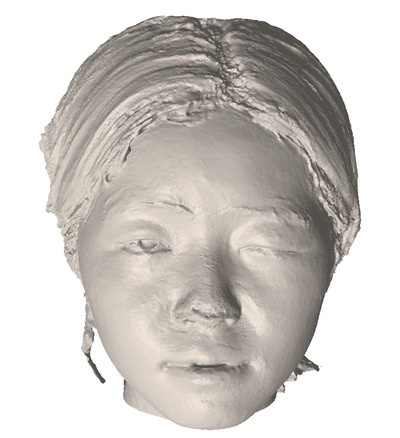} \\[2.0pt]
\includegraphics[width=\qcwbench]{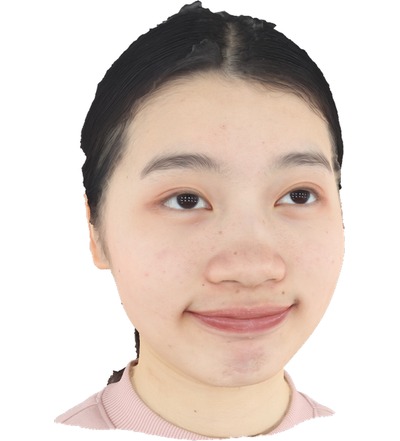} & \includegraphics[width=\qcwbench]{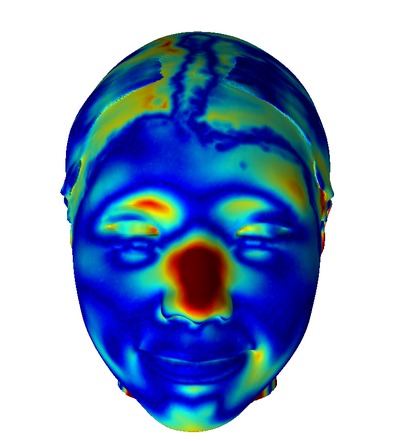} & \includegraphics[width=\qcwbench]{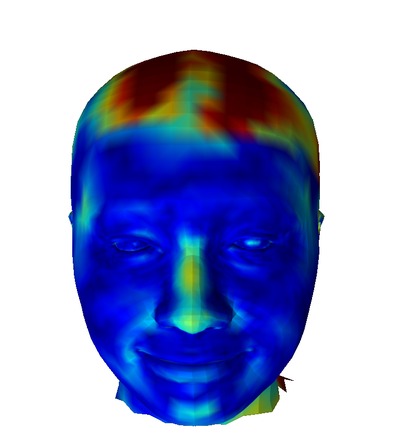} & \includegraphics[width=\qcwbench]{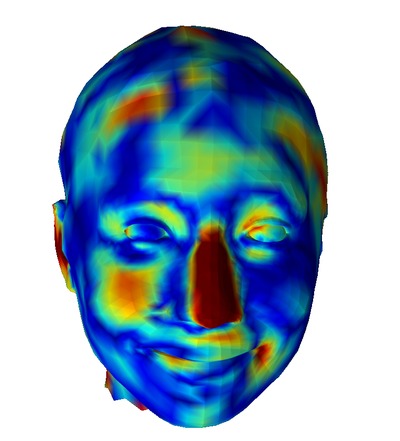} & \includegraphics[width=\qcwbench]{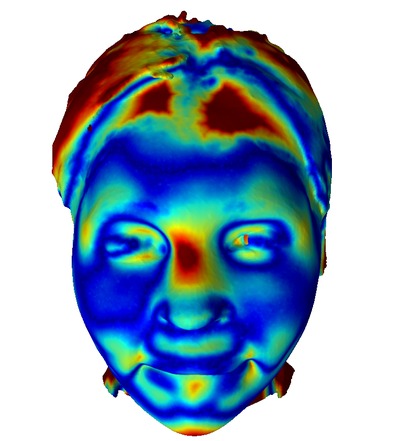} & \includegraphics[width=\qcwbench]{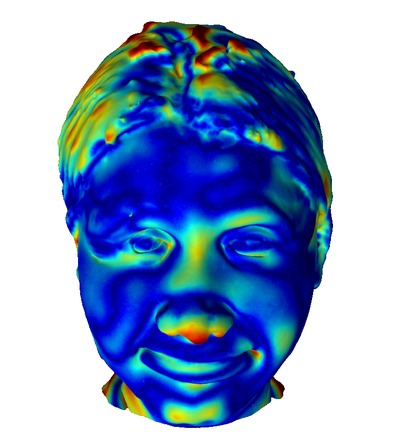} & \includegraphics[width=\qcwbench]{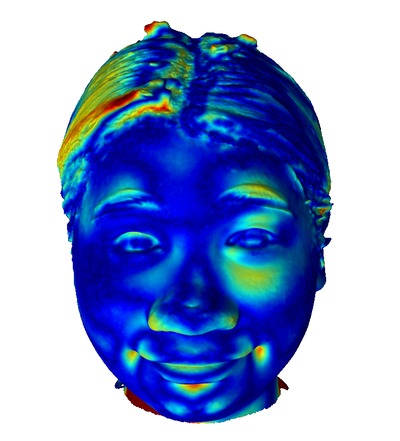} & \includegraphics[width=\qcwbench]{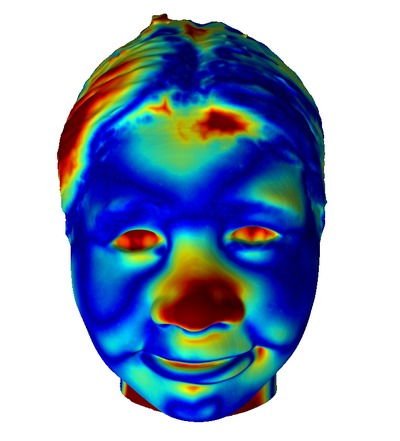} & \includegraphics[width=\qcwbench]{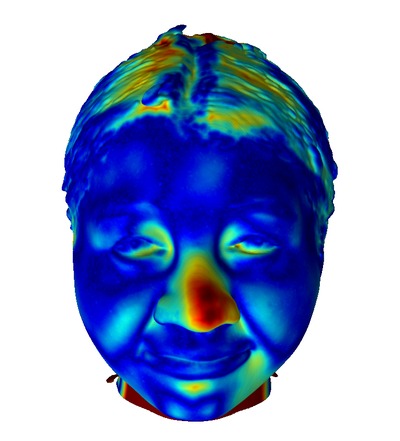} & \includegraphics[width=\qcwbench]{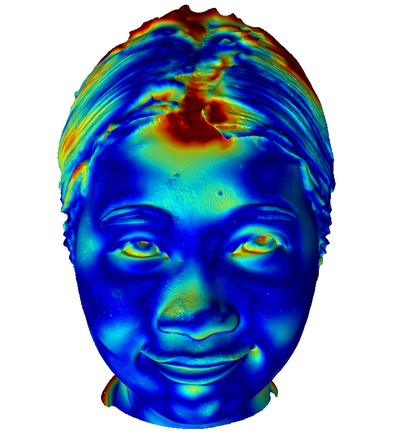} & \includegraphics[width=\qcwbench]{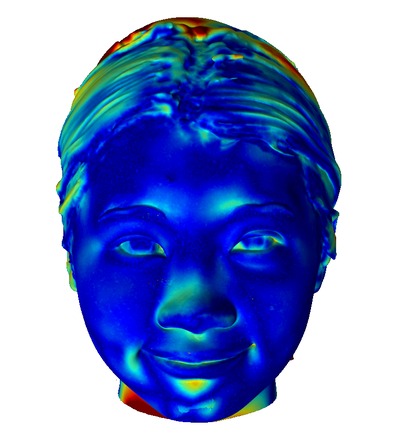} & \includegraphics[width=\qcwbench]{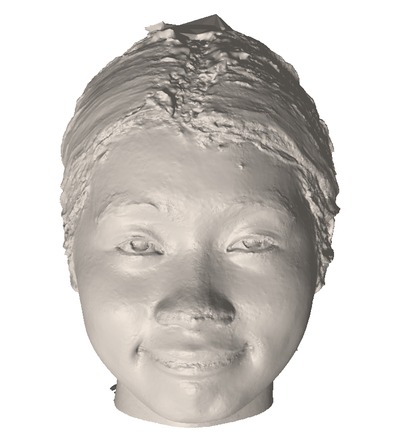} \\
\end{tabular}
}\\[2pt]
  {\scriptsize 0\,mm}\,\raisebox{-0.2ex}{\includegraphics[width=0.25\textwidth,height=0.7em]{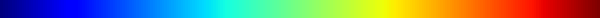}}\,{\scriptsize ${\geq}8$\,mm}
  \caption{\textbf{Mesh generation error comparison on NPHM and FaceVerse.} Colours show the distance to the GT scan.}
  \label{fig:bench}
  \vspace{8pt}
  \captionof{table}{\textbf{Mesh generation evaluation on NPHM and FaceVerse.} Head-region metrics and cross-frame consistency (Cons.), best in \textbf{bold}, second best \underline{underlined}. Parametric models share a fixed template and are consistent by construction, so their consistency is shown in gray and not ranked.}
  \label{tab:main}
  \footnotesize
  \begin{tabular*}{\linewidth}{@{\extracolsep{\fill}}lcccccccccc@{}}
\toprule
& \multicolumn{5}{c}{NPHM} & \multicolumn{5}{c}{FaceVerse} \\
\cmidrule(lr){2-6}\cmidrule(l){7-11}
Method & Chamfer$\downarrow$ & Compl.$\downarrow$ & EMD$\downarrow$ & F-score$\uparrow$ & Cons.$\downarrow$ & Chamfer$\downarrow$ & Compl.$\downarrow$ & EMD$\downarrow$ & F-score$\uparrow$ & Cons.$\downarrow$ \\
\midrule
MonoNPHM~\cite{giebenhain2024mononphm} & 8.73 & 12.57 & 27.4 & 33.0 & \textcolor{gray}{4.91} & 7.51 & 9.35 & 20.9 & 29.3 & \textcolor{gray}{5.94} \\
Pixel3DMM~\cite{giebenhain2025pixel3dmm} & 8.41 & 12.11 & 29.4 & 38.8 & \textcolor{gray}{2.07} & 9.24 & 11.81 & 27.5 & 28.3 & \textcolor{gray}{5.66} \\
VHAP~\cite{qian2024vhap} & 6.32 & 8.13 & 24.0 & 31.8 & \textcolor{gray}{4.93} & 7.95 & 9.92 & 22.9 & 25.2 & \textcolor{gray}{10.36} \\
\midrule
TRELLIS~\cite{xiang2025trellis} & 6.26 & 6.10 & 19.8 & 31.1 & 12.64 & 7.48 & 7.68 & 21.9 & 29.4 & 12.36 \\
Direct3D-S2~\cite{wu2025direct3ds2} & 6.17 & 6.03 & 21.0 & 31.8 & 11.71 & 7.65 & 6.62 & 24.3 & 29.4 & 12.07 \\
TRELLIS.2~\cite{xiang2025trellis2} & 5.99 & 5.98 & 19.3 & 34.1 & 12.37 & 5.96 & 6.57 & \underline{17.1} & 35.7 & 11.31 \\
TripoSG~\cite{li2025triposg} & 5.95 & 5.84 & 20.6 & 31.6 & 11.69 & 7.42 & 7.71 & 22.4 & 29.5 & 11.35 \\
Hunyuan3D-2.1~\cite{hunyuan2025v21} & 5.18 & 5.41 & 16.7 & 33.7 & 9.74 & 6.72 & 7.53 & 18.2 & 30.8 & 12.38 \\
Pixal3D~\cite{li2026pixal3d} & 4.83 & 5.32 & 17.3 & 38.0 & 9.48 & 6.15 & 6.71 & 17.9 & 32.9 & 11.09 \\
\midrule
\pixft & \underline{3.86} & \underline{3.77} & \underline{13.3} & \underline{45.6} & \underline{6.73} & \underline{5.14} & \underline{5.35} & 17.6 & \underline{39.4} & \underline{9.58} \\
Ours & \textbf{3.27} & \textbf{3.23} & \textbf{11.8} & \textbf{50.9} & \textbf{5.03} & \textbf{4.65} & \textbf{5.07} & \textbf{15.9} & \textbf{43.1} & \textbf{7.87} \\
\bottomrule
\end{tabular*}

\end{figure*}

\subsection{Bootstrapped Correspondences}
\label{sec:fitting}
At test time, no scan is available.
As shown in \cref{fig:pipeline} (top right), we bootstrap correspondences by fitting the head model to \pixft's per-frame outputs, whose detailed, pixel-aligned meshes provide better targets than pixels.

\paragraph{FLAME fitting.}
Fitting a parametric model to a generated mesh requires a good initial scale and pose.
Since the mesh lacks template topology, we obtain dense FLAME~\cite{li2017flame} correspondences from the image.
Pixel3DMM~\cite{giebenhain2025pixel3dmm} fits FLAME to each image.
We then project the rasterised template coordinates at face pixels onto the single-frame mesh, pairing mesh points with template coordinates.
We jointly fit a similarity $(s_f, R_f, t_f)$ and the FLAME shape, expression, and jaw parameters to these pairs using a robust point loss.
The similarity places every mesh in a canonical head frame with a consistent scale and serves as $A_f$.
For real images and our FaceVerse renders, $A_f$ combines per-frame head pose and scale, estimated from Face Anything~\cite{faceanything}, with a single FLAME fit to the first frame.
This ensures consistent placement across all frames.

\paragraph{Group fitting.}
We then fit one MonoNPHM~\cite{giebenhain2024mononphm} to all meshes of the group, with a shared identity code $z^{\mathrm{id}}$, per-frame expression codes $z_f^{\mathrm{ex}}$ and per-frame similarities $S_f$ initialised from the FLAME fit,
\begin{equation}
  \begin{aligned}
  \min_{z^{\mathrm{id}},\{z_f^{\mathrm{ex}}, S_f\}} \; &\sum_{f} \frac{1}{|P_f|}\sum_{y\in P_f} \big| \mathrm{SDF}\big(S_f(y);\, z^{\mathrm{id}}, z_f^{\mathrm{ex}}\big) \big| \\
  &+ \lambda\, \mathcal{R}\big(z^{\mathrm{id}}, \{z_f^{\mathrm{ex}}\}\big) ,
  \end{aligned}
  \label{eq:nphmfit}
\end{equation}
where $\mathrm{SDF}$ is the signed distance of MonoNPHM, $P_f$ are points sampled on the mesh of frame $f$, and $\mathcal{R}$, weighted by $\lambda$, regularises the codes towards the prior.
The shared identity lets the fit of every frame use the evidence of all frames.
MonoNPHM models each expression as a backward deformation into a shared canonical space, so $\varphi_{f\to g}$ of \cref{eq:phidef} maps a point of frame $f$ into the canonical space and from there into frame $g$.
The second step requires inverting frame $g$'s deformation, whose inverse has no closed form.
We approximate it with eight fixed-point iterations initialised at the canonical point.
The residual exceeds 1\,mm for only \ninvFailOneMm\% of points; see the appendix for details.

\paragraph{Training with fitted correspondences.}
A training group consists of scans of one identity with different expressions, each rendered from its own random viewpoint, and \cref{eq:fm} is summed over the group.
The two instances of \cref{eq:phidef} differ in kind, since registration maps are exact on the scan surface and interpolate elsewhere, whereas fitted maps follow the learned deformation field of MonoNPHM everywhere but inherit the errors of the fit.
Adapters trained on registrations alone therefore never see the kind of correspondences that they receive at test time.
We run this procedure once to precompute fitted correspondences for a fixed pool of training groups.
During training, half the groups come from this pool, while the other half use registration correspondences.
Fitted correspondences expose adapters to fitting errors, while registration groups provide accurate supervision.
Changing group size from 2 to 8 scans lets one model handle varying numbers of images.
At test time, our method first generates per-frame meshes for fitting, then jointly generates all meshes of a person.
A modified version of the standard clean-up avoids introducing holes.

\section{Experiments}
\label{sec:exp}

\subsection{Implementation Details}
\label{sec:impl}
For \pixft, we fine-tune \pix on 10{,}180 scans of the 477 NPHM identities outside our test set, each rendered from two viewpoints. S1, S2 and the cascade are trained for 8k, 8k and 12k steps, respectively, with learning rates of $10^{-5}$, $10^{-5}$ and $5{\cdot}10^{-6}$ and batch sizes of 16, 16 and 4.
For joint generation, the adapters follow every third block of both flows, with width 512, 8 heads and $K{=}4$ keys per frame in S2, which amounts to 55\,M parameters per stage.
We train them with the backbone on groups of 2 to 8 scans from the 459 of these 477 identities with at least four scans.
Each stage is trained for 4k steps with two groups per step in S1 and one in S2 and learning rates of $10^{-4}$ for the adapters and $10^{-5}$ for the backbone.
Both trainings use AdamW, 200 warm-up steps and a condition dropout of 0.1 for classifier-free guidance.
On NVIDIA A100 GPUs, the fine-tuning uses two GPUs per stage, and the adapter training takes about 7\,h for S1 and 6\,h for S2 on one GPU.
For eight images, our method takes about \nRtsFactor times as long as generating them one by one with \pixft.

\subsection{Evaluation Protocol}
\label{sec:setup}
\paragraph{Scan benchmarks.}
We evaluate on 10 held-out NPHM~\cite{giebenhain2023nphm} identities, each with eight expressions.
Each scan is rendered photorealistically from a random viewpoint.
We also evaluate on 10 FaceVerse~\cite{wang2022faceverse} identities, each with eight scans. These scans differ from our training data. We render them from varied viewpoints and process them like real images.

\paragraph{Real images.}
We evaluate on 10 NeRSemble~\cite{kirschstein2023nersemble} identities outside the NPHM training and test sets and 10 AVA-256~\cite{martinez2024ava256} identities.
Each person's eight input images show four expressions captured by two synchronised cameras, one on each side of the face. The remaining cameras provide held-out views and multi-view stereo points from COLMAP~\cite{schoenberger2016sfm,schoenberger2016mvs}.
Video frames from VFHQ~\cite{xie2022vfhq} and CelebV-HQ~\cite{zhu2022celebvhq} serve for qualitative results, and all methods receive the same RGBA crops.

\paragraph{Baselines.}
We compare against the parametric trackers VHAP~\cite{qian2024vhap} and Pixel3DMM~\cite{giebenhain2025pixel3dmm}, which fit a shared identity across a person's images.
We also evaluate MonoNPHM~\cite{giebenhain2024mononphm} and six image-to-3D generators~\cite{xiang2025trellis,wu2025direct3ds2,xiang2025trellis2,li2025triposg,hunyuan2025v21,li2026pixal3d} on each image independently.
Finally, \pixft uses our fine-tuned backbone to generate a mesh independently for each image.
Our method and \pixft use only the input camera, as described in \cref{sec:fitting}.
Ground-truth geometry and all other cameras are used only for evaluation.

\begin{figure}[t]
  \centering
  {\setlength{\tabcolsep}{1.0pt}\renewcommand{\arraystretch}{0}
\newlength{\bhcw}\setlength{\bhcw}{\dimexpr(\linewidth-1.2em-8pt)/4\relax}
\begin{tabular}{@{}c@{\hspace{2pt}}cccc@{}}
\rotatebox{90}{\makebox[1.0508\bhcw][c]{\fontsize{6.2pt}{7.2pt}\selectfont Input}} & \includegraphics[width=\bhcw,height=1.0508\bhcw]{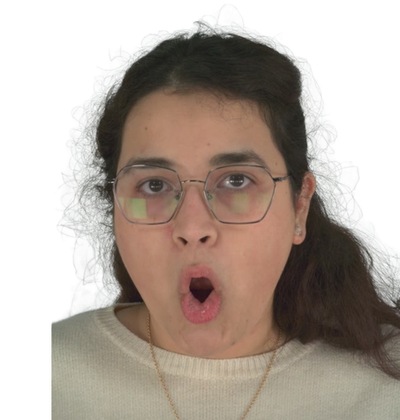} & \includegraphics[width=\bhcw,height=1.0508\bhcw]{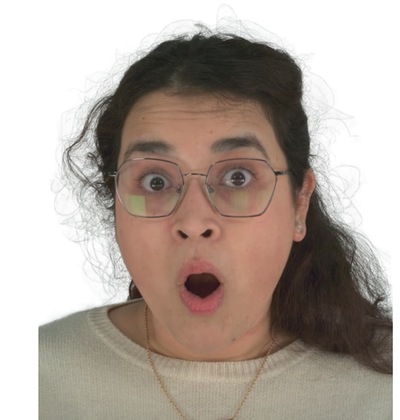} & \includegraphics[width=\bhcw,height=1.0508\bhcw]{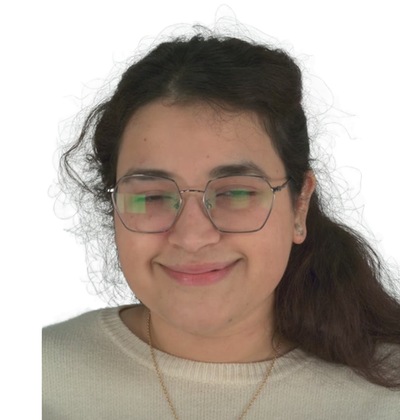} & \includegraphics[width=\bhcw,height=1.0508\bhcw]{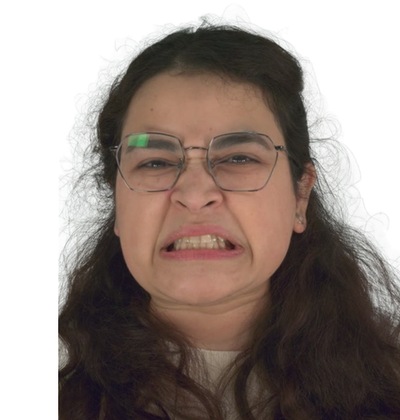} \\[1.5pt]
\rotatebox{90}{\makebox[1.0508\bhcw][c]{\fontsize{6.2pt}{7.2pt}\selectfont \pixft}} & \includegraphics[width=\bhcw,height=1.0508\bhcw]{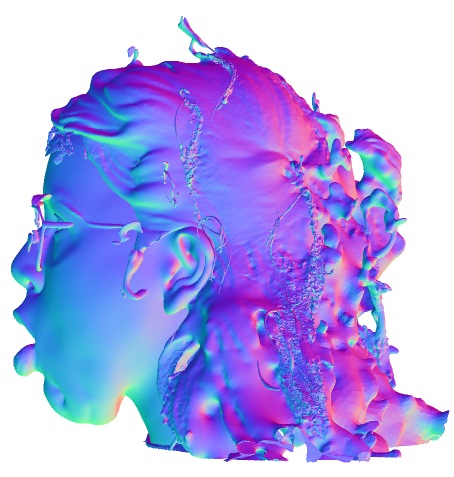} & \includegraphics[width=\bhcw,height=1.0508\bhcw]{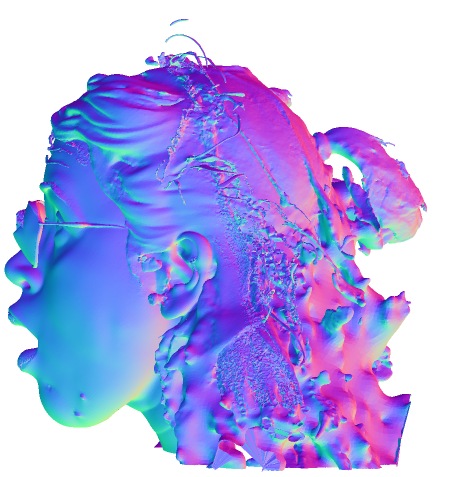} & \includegraphics[width=\bhcw,height=1.0508\bhcw]{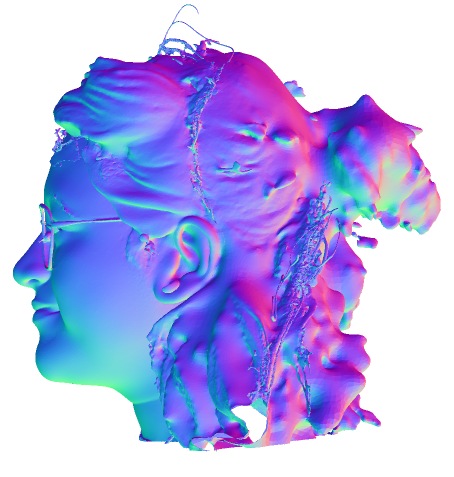} & \includegraphics[width=\bhcw,height=1.0508\bhcw]{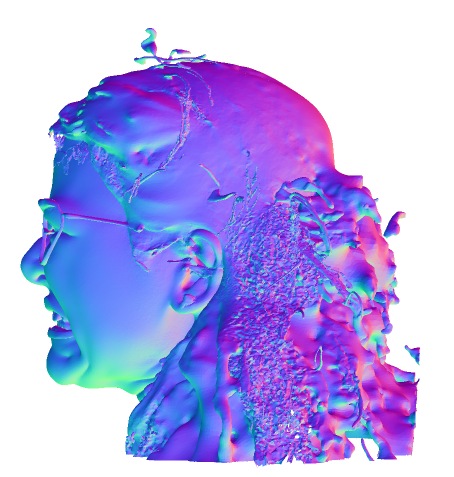} \\[1.5pt]
\rotatebox{90}{\makebox[1.0508\bhcw][c]{\fontsize{6.2pt}{7.2pt}\selectfont Ours}} & \includegraphics[width=\bhcw,height=1.0508\bhcw]{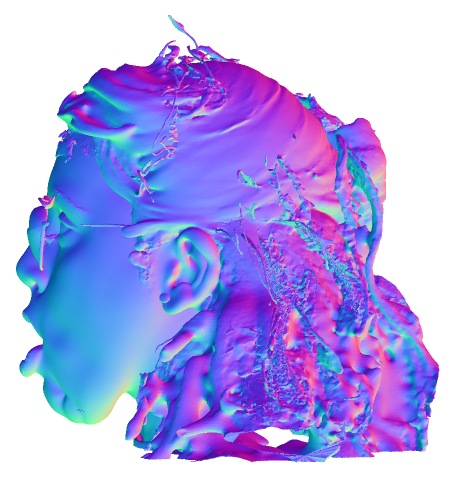} & \includegraphics[width=\bhcw,height=1.0508\bhcw]{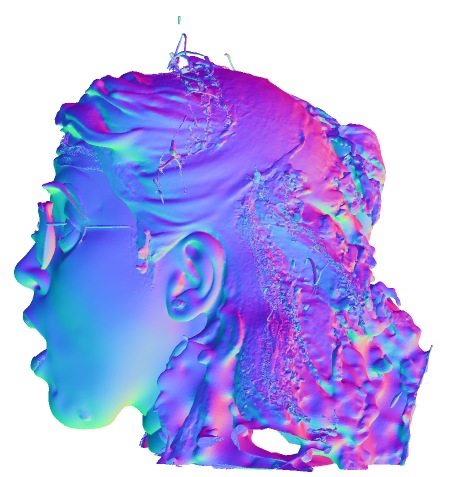} & \includegraphics[width=\bhcw,height=1.0508\bhcw]{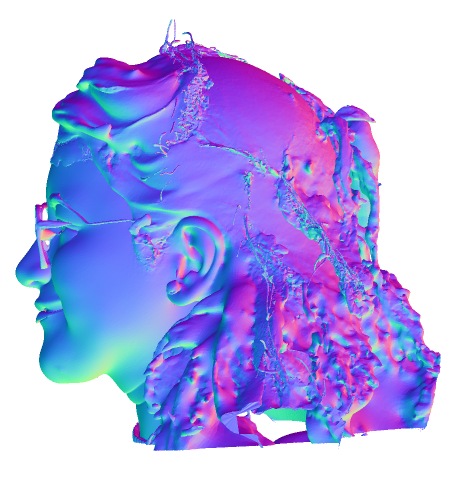} & \includegraphics[width=\bhcw,height=1.0508\bhcw]{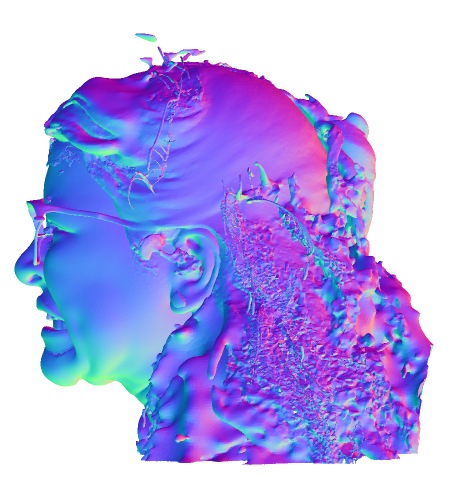} \\[1.5pt]
\end{tabular}
}
  \caption{\textbf{Back-of-head consistency on NeRSemble.} Side views of the meshes generated from four input frames. \pixft hallucinates a different hairstyle for every frame, while ours agree.}
  \label{fig:backhead}
\end{figure}

\paragraph{Metrics.}
Every prediction, ours included, is aligned to its scan with the same similarity search~\cite{besl1992icp,umeyama1991} and a final rigid face alignment, so that the metrics compare shapes rather than placements.
In the head region above a plane 2\,cm below the chin, we report the Chamfer distance, the completeness, the Earth Mover's distance and the F-score at 2\,mm~\cite{knapitsch2017tanks}.
We warp each aligned prediction's non-face surface into every other frame using ground-truth deformation.
Cross-frame consistency measures the mean distance to the target predictions.
For the captures, we report silhouette IoU in held-out cameras after a similarity fit using the alignment cameras.
We also report the Chamfer distance between the camera-visible head and stereo points.
We measure consistency by the head Chamfer distance between meshes generated from synchronised images of the same instant.
All distances are in mm and F-scores in \%.

\subsection{Comparison with the State of the Art}
\label{sec:results}
\paragraph{Qualitative comparison.}
\Cref{fig:real} compares meshes for frames of NeRSemble, AVA-256, VFHQ and CelebV-HQ.
The parametric models recover plausible faces but represent hair as a smooth volume at best, while image-to-3D generators often fail to capture specific facial shape and produce noisy hair geometry.
Our meshes have a higher surface quality and more detail, both in facial features such as the eyes, lips and wrinkles and in the structure of the hair.

\paragraph{Scan benchmarks.}
\Cref{tab:main} evaluates all methods against the ground-truth scans.
Our method is the most accurate on both datasets, including FaceVerse, whose scans and inputs differ from our training data, whereas the parametric models produce coarse heads without hair.
Their shared template makes them consistent by construction, which is why we do not rank them in consistency.
Fine-tuning on head scans already makes \pixft the strongest generator, and joint generation improves over it in accuracy and clearly in cross-frame consistency, although both share one backbone.
The error maps in \cref{fig:bench} confirm that our meshes are closer to the scans across the whole head.

\begin{table}[t]
  \centering
  \caption{\textbf{Mesh generation evaluation on NeRSemble and AVA-256.} Inputs are real images from two synchronised cameras. Held-out IoU, Chamfer (Ch.) and cross-view consistency (Cons.); gray values are not ranked.}
  \label{tab:mv10}
  \footnotesize\setlength{\tabcolsep}{2pt}
  \begin{tabular*}{\linewidth}{@{\extracolsep{\fill}}lcccccc@{}}
\toprule
& \multicolumn{3}{c}{NeRSemble} & \multicolumn{3}{c}{AVA-256} \\
\cmidrule(lr){2-4}\cmidrule(l){5-7}
Method & IoU$\uparrow$ & Ch.$\downarrow$ & Cons.$\downarrow$ & IoU$\uparrow$ & Ch.$\downarrow$ & Cons.$\downarrow$ \\
\midrule
MonoNPHM~\cite{giebenhain2024mononphm} & 0.902 & 5.07 & \textcolor{gray}{4.03} & 0.869 & 4.74 & \textcolor{gray}{3.93} \\
Pixel3DMM~\cite{giebenhain2025pixel3dmm} & 0.908 & 4.50 & \textcolor{gray}{1.36} & 0.884 & 4.80 & \textcolor{gray}{1.77} \\
VHAP~\cite{qian2024vhap} & 0.932 & 4.20 & \textcolor{gray}{2.70} & 0.882 & 4.31 & \textcolor{gray}{3.22} \\
\midrule
TRELLIS~\cite{xiang2025trellis} & 0.923 & 3.93 & 5.70 & 0.900 & 3.92 & 5.49 \\
Direct3D-S2~\cite{wu2025direct3ds2} & 0.931 & 4.80 & 7.86 & \underline{0.911} & 4.49 & 7.04 \\
TRELLIS.2~\cite{xiang2025trellis2} & 0.922 & 4.06 & 6.09 & 0.870 & 4.23 & 6.76 \\
TripoSG~\cite{li2025triposg} & 0.936 & 3.88 & 5.31 & 0.902 & 3.96 & 5.30 \\
Hunyuan3D-2.1~\cite{hunyuan2025v21} & 0.923 & 4.65 & 8.35 & 0.899 & 3.96 & 5.50 \\
Pixal3D~\cite{li2026pixal3d} & 0.931 & 3.88 & 5.36 & 0.900 & 3.46 & 4.80 \\
\midrule
\pixft & \underline{0.942} & \underline{3.29} & \underline{4.37} & 0.905 & \underline{3.36} & \underline{4.22} \\
Ours & \textbf{0.953} & \textbf{3.10} & \textbf{3.30} & \textbf{0.915} & \textbf{3.13} & \textbf{3.12} \\
\bottomrule
\end{tabular*}

\end{table}

\paragraph{Real camera images.}
As \cref{tab:mv10} shows, our method outperforms all baselines on the real images of NeRSemble and AVA-256.
Our method produces more accurate held-out silhouettes, indicating better reconstruction of unseen regions.
Its visible head surfaces are closer to the stereo points, and meshes from synchronised cameras agree more closely than those of every other generator.
Joint generation improves all three metrics over \pixft, which shares our backbone, on both datasets.

\paragraph{Consistency of unseen regions.}
\Cref{fig:backhead} shows the back of the head, which none of the inputs shows.
\pixft hallucinates a different hairstyle when generating each frame independently, whereas our meshes, denoised jointly and coupled at corresponding points, share one hairstyle.

\begin{table}[t]
  \centering
  \caption{\textbf{FLAME tracking accuracy on AVA-256.} Each ``+'' row adds a 3D term towards a per-frame prediction to the Pixel3DMM fit.}
  \label{tab:flametrack}
  \footnotesize
  \begin{tabular*}{\linewidth}{@{\extracolsep{\fill}}lcc@{}}
\toprule
FLAME tracker & Chamfer$\downarrow$ & F-score$\uparrow$ \\
\midrule
Pixel3DMM~\cite{giebenhain2025pixel3dmm} & 4.86 & 46.1 \\
\quad + Face Anything~\cite{faceanything} & \underline{4.44} & \underline{51.9} \\
\quad + Ours & \textbf{4.26} & \textbf{52.4} \\
\bottomrule
\end{tabular*}

\end{table}

\paragraph{Downstream FLAME tracking.}
Our meshes can also constrain a tracker.
On frontal-camera AVA-256 images, we refine Pixel3DMM's~\cite{giebenhain2025pixel3dmm} multi-image FLAME fit with a robust point-to-surface term using our meshes or Face Anything's~\cite{faceanything} per-pixel geometry.
Both use identical placement and weighting, without ground truth.
As \cref{tab:flametrack} shows, both constraints improve the tracker, and our meshes, which cover the whole head, improve it most.

\begin{figure}[t]
  \centering
  \includegraphics[width=\linewidth]{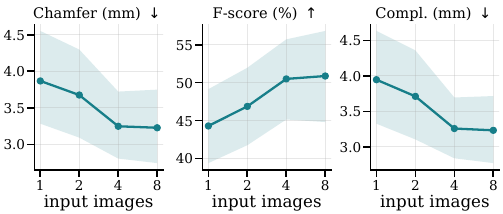}
  \caption{\textbf{Number of images.} Head-region metrics of a target frame generated jointly with 1, 2, 4 or 8 images.}
  \label{fig:nsweep}
\end{figure}

\paragraph{Number of images.}
\Cref{fig:nsweep} gives the mean and 95\% bootstrap interval over \nnsTargets NPHM targets. With fixed noise, the target mesh becomes more accurate and complete as images are added, with the largest gains from the first few.

\subsection{Ablation Study}
\label{sec:ablation}
\Cref{tab:ablation} evaluates the ablations on NPHM with the metrics of \cref{tab:main}. Each variant changes one component and uses the same test-time fits.
Synchronising \pixft without training, by averaging the predicted occupancy of corresponding voxels at every step~\cite{bartal2023multidiffusion,kim2024synctweedies}, forces the frames to agree and yields the best consistency. However, consistency measures agreement rather than fidelity. Averaging erodes the heads and yields the largest errors, supporting learned cross-frame information exchange.
Global attention over all tokens of other frames is clearly less accurate, as it does not learn where to attend from a few hundred identities.
We test the correspondence map against the GT-only variant, trained on ground-truth correspondences only, whose adapters never see the noisy fitted correspondences of test time.
Same-voxel attention, as in video models, attends to the same voxel in the other frames and falls clearly behind, since that voxel holds a different point of the head.
FLAME correspondences attend to the points that fitted FLAME surfaces match; the more accurate NPHM correspondences of the GT-only variant generate better heads, with lower errors and a clearly higher F-score.
Finally, mixing in fitted correspondences, which teaches the adapters to cope with their errors, improves every metric over the GT-only variant.

\begin{table}[t]
  \centering
  \caption{\textbf{Ablation study.} Each variant changes one component of the row it is compared with; the full model, which combines all components, is best in every accuracy metric.}
  \label{tab:ablation}
  \footnotesize\setlength{\tabcolsep}{1.6pt}
  \begin{tabular*}{\linewidth}{@{\extracolsep{\fill}}lccccc@{}}
\toprule
Variant & Chamfer$\downarrow$ & Compl.$\downarrow$ & EMD$\downarrow$ & F-score$\uparrow$ & Cons.$\downarrow$ \\
\midrule
Training-free sync & 7.61 & 12.38 & 35.8 & 42.1 & \textbf{4.25} \\
Global attention & 3.99 & 3.78 & 13.8 & 45.1 & 6.54 \\
Same-voxel attention & 3.94 & 3.80 & 13.7 & 46.1 & 6.31 \\
FLAME corresp. & 3.60 & 3.50 & 12.4 & 47.8 & 5.64 \\
GT-only training & \underline{3.45} & \underline{3.38} & \underline{12.0} & \underline{50.5} & 5.11 \\
\midrule
Ours & \textbf{3.27} & \textbf{3.23} & \textbf{11.8} & \textbf{50.9} & \underline{5.03} \\
\bottomrule
\end{tabular*}

\end{table}

\section{Limitations and Future Work}
\label{sec:limitations}
Despite its strong performance, our method is specialised for human heads because it derives correspondences from a parametric head model.
It therefore does not directly generalise to full human bodies or generic objects.
Extending correspondence-routed joint generation to these settings, for example through body models or category-level correspondences, is a promising direction.
The underlying parametric fit becomes less reliable under heavy occlusion or when only a small portion of the face is visible.
Our method also assumes that head geometry remains consistent across images, apart from expression changes.
Photo collections taken at different times, with changes in hairstyle or accessories, therefore fall outside its scope.
Finally, its reliance on a computationally expensive diffusion-based image-to-mesh generator makes real-time applications infeasible.

\section{Conclusion}
\label{sec:conclusion}
\ours turns a pretrained single-image 3D generator into a consistent joint generator for any number of images of a person.
Each 3D token attends to corresponding points in other frames.
We obtain these correspondences by fitting a parametric head model to the generator's outputs, requiring no ground truth at test time.
Training on both ground-truth and fitted correspondences makes the adapters robust to fitting errors.
Each mesh follows its own image, while regions unseen in that image are completed using information from all images.
On scan benchmarks and real captures, \ours improves accuracy over generator and parametric baselines.
It also improves consistency over image-to-3D generators, while its meshes improve parametric face tracking.
More broadly, correspondence-routed attention lets a single-image 3D generator exploit several images without training a multi-image prior from scratch.

\section*{Acknowledgments}
This work was supported by the ERC Consolidator Grant Gen3D (101171131) of Matthias Nie{\ss}ner. We thank Angela Dai for the video voice-over.

\section*{LLM Usage Statement}
We used Claude (Anthropic), mainly through the Claude Code coding assistant, to support this work. Under the authors' direction, it assisted with implementing and running much of the code, including the training and evaluation pipelines, baseline runs, metric and scoring scripts, and job management on our compute cluster. It also assisted with scripts that generate the tables, plots and figure layouts, including the TikZ sources of \cref{fig:pipeline,fig:adapter}. The authors specified the experiments, inspected the results and checked the reported numbers. It supported discussions of research directions and the analysis of experimental results. The authors chose the research direction, the method and all design decisions. It helped draft and edit the text and verify the citations. All text was revised and approved by the authors, who take full responsibility for the entire content of this paper.

{\small
\bibliographystyle{ieeenat_fullname}
\bibliography{references}
}

\clearpage
\twocolumn[{\centering\Large\textbf{Appendix}\par\vspace{1.0em}}]
\appendix
\renewcommand{\theHtable}{S\arabic{table}}\renewcommand{\theHfigure}{S\arabic{figure}}\renewcommand{\theHequation}{S\arabic{equation}}
\setcounter{table}{0}\renewcommand{\thetable}{S\arabic{table}}
\setcounter{figure}{0}\renewcommand{\thefigure}{S\arabic{figure}}
\setcounter{equation}{0}\renewcommand{\theequation}{S\arabic{equation}}

\section{Overview}
\label{sec:supp_overview}
This appendix complements the main paper.
\Cref{sec:supp_video} describes our supplementary video, \cref{sec:supp_impl} gives additional implementation details, \cref{sec:supp_protocol} describes the benchmarks and the evaluation protocol, and \cref{sec:supp_full} reports additional comparisons.
\Cref{sec:supp_ablation} adds ablations, \cref{sec:supp_runtime} reports the runtime and memory of every step, and \cref{sec:supp_qual} shows further results of our method.

\FloatBarrier
\section{Supplementary Video}
\label{sec:supp_video}
We highly recommend watching our supplementary video, a narrated presentation of our method that contains additional qualitative results.
The video shows renderings of our generated head meshes, in which an orbiting camera turns around each mesh and reveals all of its details, including the hair, the ears and the back of the head.
It further compares our meshes with those of the baselines, showing the reconstruction quality and fidelity of each method as well as the consistency of the meshes generated from different images of the same person.

\FloatBarrier
\section{Additional Implementation Details}
\label{sec:supp_impl}

\subsection{Base Model and Sampling}
We build on the public code and the released weights of \pix~\cite{li2026pixal3d}, which follows the two-stage design of TRELLIS.2~\cite{xiang2025trellis2} with DINOv3~\cite{simeoni2025dinov3} conditioning, and sample all stages with its released sampler settings.
The noise of every stage is seeded per frame, so a frame receives the same noise in the single-image pass and in every joint group.
The cascade from resolution $512^3$ to $1024^3$ and the decoding run per frame on the jointly generated S2 latent.

\subsection{Single-Image Fine-Tuning}
\label{sec:supp_finetune}
\pixft is \pix fine-tuned in all three geometry stages with \cref{eq:fm} and no architectural change.
It is fine-tuned on renders of the scans, and our adapters are trained on the same renders.
We rasterise each scan with its captured vertex colours from random viewpoints with varying fields of view.
We match foreground colour statistics to real video frames and apply slight Gaussian blur, sensor noise and a JPEG round trip.
Finally, we crop around the head as in the real-image pipeline.
Among several variants of this renderer, this one was closest to real photographs in DINOv3 feature statistics.

\subsection{Adapters and Joint Training}
\paragraph{Architecture.}
The adapters follow every third block of both flows, starting with the third block.
Each adapter uses LayerNorm, width-512 query, key and value projections with 8 heads, the key embeddings of \cref{eq:keys}, and a learned null key-value pair.
A zero-initialised projection maps back to the backbone width, with its output scaled by a gate on the timestep embedding.
The key embeddings combine a depth MLP, used in both stages, with a validity embedding in S1 and an offset MLP in S2, and every stage has 55\,M adapter parameters.
The depth features $d_i$ and $d_j$ are the coordinates of the query and the key token along the viewing axis of the view-aligned latent grid of their own frame.

\paragraph{Training.}
Registration groups and fitted groups are drawn with equal probability.
All frames of a group share the flow time, which S1 draws from a logit-normal distribution.
The backbone is fine-tuned together with the adapters, with a learning rate ten times lower than that of the adapters.

\subsection{Correspondences}
\label{sec:supp_corr}
\paragraph{Registration correspondences.}
For registration groups, $\varphi_{f\to g}$ in \cref{eq:phi} uses the 8 nearest registration vertices with inverse-distance weights.
$A_f$ is the known similarity transform from the scan to the render's view-aligned latent grid.
In S1, a correspondence is valid if $\Phifg(p)$ lies inside the latent grid; in S2, a key token is valid if it lies within two tokens of $\Phifg(p)$.
Invalid keys are masked, and the null key is always available.

\paragraph{FLAME fitting.}
The test-time FLAME fit minimises
\begin{equation}
  \min_{s_f,R_f,t_f,\beta_f} \sum_{(y,\hat{c})\in\mathcal{C}_f} \rho\big( s_f R_f\, F_{\beta_f}(\hat{c}) + t_f - y \big) + \lambda_\beta \|\beta_f\|^2 ,
  \label{eq:flamefit}
\end{equation}
over the pairs $\mathcal{C}_f$ of a mesh point $y$ and a FLAME template coordinate $\hat{c}$, where $F_{\beta_f}(\hat{c})$ is the posed FLAME surface point at $\hat{c}$, $\rho$ a Huber penalty and $\lambda_\beta$ a prior weight on the shape, expression and jaw parameters $\beta_f$ of FLAME 2023~\cite{li2017flame}.
The similarity is initialised by a Procrustes alignment.
The pairs come from face, eye, nose, lips and forehead pixels of the Pixel3DMM~\cite{giebenhain2025pixel3dmm} fit whose rays intersect the single-frame mesh in the generator's camera.
We do not use the fit's depth or scale.

\paragraph{MonoNPHM group fit.}
We optimise \cref{eq:nphmfit} with Adam.
The identity code starts at the mean of the identity codes of MonoNPHM~\cite{giebenhain2024mononphm} and the expression codes at zero, and $\mathcal{R}$ penalises both codes after normalising them by the standard deviations of the code banks of MonoNPHM.
Every iteration uses random surface points of the single-frame meshes above the neck.
The fitted pool of the training groups was computed once with the same procedure, from single-frame meshes of the backbone after its first fine-tuning stage.

\paragraph{Inverse deformation.}
$\varphi_{f\to g}$ maps $y$ to $\bar{y} = S_f(y) + \mathcal{D}(S_f(y); z_f^{\mathrm{ex}})$ using MonoNPHM's backward deformation $\mathcal{D}$, then returns $S_g^{-1}(\tilde{y})$.
We solve $\tilde{y} + \mathcal{D}(\tilde{y}; z_g^{\mathrm{ex}}) = \bar{y}$ approximately with eight iterations of $\tilde{y} \leftarrow \bar{y} - \mathcal{D}(\tilde{y}; z_g^{\mathrm{ex}})$, starting from $\tilde{y} = \bar{y}$.
Deformation blends to zero on the rigid torso.
Correspondences outside the grid or without nearby tokens are masked; the null key lets tokens ignore incorrect matches.
On the test-time fits of the 10 NPHM groups, the final residual has a median \ninvResMedTxt\,mm and exceeds 1\,mm for \ninvFailOneMm\% of the points, and \ninvOutGrid\% of the S1 correspondences leave the latent grid.

\paragraph{Relation of the two kinds of correspondences.}
Both instances of \cref{eq:phidef} map a point of the head in one frame to the same point in another frame, but they are built differently.
Registration maps interpolate the displacements of corresponding vertices, so they are exact on the scan surface and extrapolate off the surface by inverse-distance weighting.
Fitted maps evaluate MonoNPHM's learned deformation field, which is smooth and defined everywhere, but inherit errors from the FLAME and MonoNPHM fits.
Their placement $A_f$ comes from the FLAME fit rather than the scan.
Training on registrations alone would leave a gap between training and test-time correspondences, which the fitted pool closes.
The GT-only variant in \cref{tab:ablation} and the fitted-only variant in \cref{tab:supp_abl_design} evaluate the two extremes.

\subsection{Clean-Up}
\label{sec:supp_holes}
The pipeline's standard mesh clean-up closes small holes, cuts the decoded surface along non-manifold edges and removes connected components below an area threshold.
Removing pieces disconnected from the head by these cuts opens holes, most visibly around the mouth.
Our clean-up applies the same repairs but removes floating components only after re-welding the cut, then fills small loops.
Above the head region's lower boundary, it closes each remaining boundary loop of perimeter 10--250\,mm with a flat ring patch over the hole border.
For real images, the minimum perimeter is 3\,mm.
This removes most holes and leaves the geometry metrics essentially unchanged (\cref{tab:holes}).
We apply this clean-up to both \ours and \pixft meshes.
It is not a contribution of this work but makes the meshes usable as hole-free head surfaces.
\begin{table}[t]
  \centering
  \caption{\textbf{Holes on NPHM.} Mean number of open boundaries per frame in the head region and the head metrics of \cref{tab:main} for the meshes of \ours.}
  \label{tab:holes}
  \footnotesize
  \begin{tabular}{@{}lccc@{}}
\toprule
Clean-up & Holes$\downarrow$ & Chamfer$\downarrow$ & F-score$\uparrow$ \\
\midrule
Standard clean-up & 63.2 & 3.26 & 50.9 \\
Topology-preserving + filling & 4.2 & 3.27 & 50.9 \\
\bottomrule
\end{tabular}

\end{table}

\subsection{Real-Image Pipeline}
\label{sec:supp_realimg}
For a sequence of real images, as for the FaceVerse renders, all frames are cropped with one square crop around the matte of the first frame, enlarged by 10\%.
This crop defines one virtual camera in which every frame is generated.
Its intrinsics come from capture calibration, the FaceVerse render camera or, for in-the-wild images, the focal length predicted by Face Anything~\cite{faceanything}.
Face Anything predicts metric depth and canonical coordinates for each frame, from which we obtain its rigid head motion relative to this camera.
Scaled ray-ICP against the predicted depth then estimates the metric scale of each single-frame mesh.
On NPHM, every frame is generated in its own render camera, and $A_f$ comes from per-frame FLAME fits.

\FloatBarrier
\section{Benchmarks and Evaluation Protocol}
\label{sec:supp_protocol}

\subsection{Datasets}
\paragraph{NPHM.}
We use the 10 held-out NPHM~\cite{giebenhain2023nphm} identities 088, 127, 195, 243, 291, 339, 387, 436, 482 and 535, which are not used in training, each with eight expressions.
They are rendered with the photorealistic renderer of \cref{sec:supp_finetune} from random viewpoints with a yaw within $\pm30^\circ$, a pitch between $-8^\circ$ and $20^\circ$ and a field of view between $10^\circ$ and $70^\circ$.
Each group of eight frames is generated jointly, and all 80 frames are scored.

\paragraph{FaceVerse.}
We randomly select 10 FaceVerse~\cite{wang2022faceverse} identities with artifact-free scans, each with eight expressions.
Every scan is rendered from a random viewpoint with a yaw within $\pm30^\circ$ and a pitch between $-8^\circ$ and $20^\circ$, with one camera distance per identity and a field of view at which the head fits into the image in all eight frames.
Our pipeline generates all eight images in one virtual camera (\cref{sec:supp_realimg}), so the change of viewpoint appears as head motion.
Each ground-truth scan is cut 20\,mm below the chin.
Since the scans are not metric, a single dataset-wide scale factor, shared by all methods, sets the median outer eye-corner width to 90\,mm.

\paragraph{NeRSemble and AVA-256.}
We use the 10 NeRSemble~\cite{kirschstein2023nersemble} identities 139, 170, 171, 175, 176, 177, 178, 192, 497 and 510, none of which is an NPHM training or test identity, and the 10 AVA-256~\cite{martinez2024ava256} identities AAN112, ADL311, AEY864, AJR151, ANX726, CDR970, DOT682, FXN596, KDA058 and NRE683.
For each identity, two synchronised cameras about $20^\circ$ to $33^\circ$ to the left and right of the face observe four timestamps: a near-neutral face, a smile, an open mouth and one further expression, where a cheek expression replaces the smile on AVA-256, which has no smile sequence.
The two cameras are chosen by a fixed rule among those that see the whole head at all four timestamps, and NeRSemble timestamps in which the head leaves one of the two views are not used.
Since the methods assume one intrinsic matrix per sequence, the AVA-256 images of the two cameras are resampled into one virtual pinhole camera per identity with the real camera rotations and centres.
For evaluation, we hold out all other cameras of an identity, seven on NeRSemble and seven or eight on AVA-256, and reconstruct the multi-view stereo points of each timestamp with COLMAP~\cite{schoenberger2016sfm,schoenberger2016mvs} from the calibrated cameras.

\paragraph{Inputs.}
All methods receive the same RGBA crops, with BiRefNet~\cite{zheng2024birefnet} mattes in the wild and the dataset mattes on the captures.
We binarise mattes for every method because soft edges cause some generators to produce flat cards.
Parametric trackers receive the same crops composited on white, as their pipelines require.
The image-to-3D generators are run with their official code and default settings on every image.

\subsection{Alignment and Metrics}
\paragraph{Scan benchmarks.}
Every prediction is aligned to its scan using the same similarity search~\cite{besl1992icp,umeyama1991}.
We run scaled trimmed ICP from 72 starting orientations, using a two-sided residual to prevent scale collapse.
We then run face ICP from each resulting pose, followed by 12 perturbed restarts from the best poses.
We keep the pose with the lowest face residual and finally align the face rigidly, so that the metrics compare shapes rather than placements.
The face and lips regions are defined by a FLAME fit of each scan, and the head region by a plane 2\,cm below the chin of that fit.
To measure cross-frame consistency, we move each aligned prediction's non-face points into every other frame of the same person using the inverse-distance-weighted displacement of the 8 nearest correspondence vertices.
Correspondences come from NPHM registrations or FLAME fits to FaceVerse scans.
We measure the mean distance to each target frame's aligned non-face surface and average over all target frames.

\paragraph{Captures.}
For silhouette IoU, we fit a per-frame similarity transform to masks from the alignment cameras, including the input cameras, by maximising a differentiable soft-silhouette IoU from 144 starts.
We then measure IoU in held-out cameras within a head region shared by all methods.
For the Chamfer distance, every mesh is aligned to the stereo points of its timestamp with the similarity search above, and only the prediction vertices within 10\,mm of a camera-visible stereo point are used.
The cross-view consistency is the head Chamfer distance between the two meshes generated from the two synchronised images of one timestamp, each placed by its own alignment to the stereo points.

\FloatBarrier
\section{Additional Quantitative Results}
\label{sec:supp_full}
\paragraph{TRELLIS with all images.}
TRELLIS~\cite{xiang2025trellis} can also be conditioned on all eight images of a person at once, in its multi-image mode with multidiffusion~\cite{bartal2023multidiffusion}, which outputs one mesh per person.
\Cref{tab:supp_trellis_mi} compares this mode with TRELLIS applied to every image on its own and with our method on NPHM, with the metrics of \cref{tab:main}.
The additional images make TRELLIS more accurate, but our meshes are far more accurate in every metric.
The multi-image mode shares one mesh per person, making it consistent by construction, like the parametric models in \cref{tab:main}.
However, it cannot follow per-frame expressions, so its consistency is shown in gray and not ranked.
\begin{table}[t]
  \centering
  \caption{\textbf{TRELLIS with all images of a person on NPHM.} The multi-image mode receives all eight images and gives one mesh per person, which is scored on every frame with the protocol of \cref{tab:main}. Distances in mm and F-score in \%; the consistency of the shared multi-image mesh is shown in gray and not ranked.}
  \label{tab:supp_trellis_mi}
  \footnotesize
  \begin{adjustbox}{max width=\linewidth}\begin{tabular}{@{}lccccc@{}}
\toprule
Method & Chamfer$\downarrow$ & Compl.$\downarrow$ & EMD$\downarrow$ & F-score$\uparrow$ & Cons.$\downarrow$ \\
\midrule
TRELLIS~\cite{xiang2025trellis} & 6.26 & 6.10 & 19.8 & 31.1 & \underline{12.64} \\
TRELLIS (multi-image) & \underline{5.58} & \underline{5.47} & \underline{19.1} & \underline{34.4} & \textcolor{gray}{4.88} \\
\midrule
Ours & \textbf{3.27} & \textbf{3.23} & \textbf{11.8} & \textbf{50.9} & \textbf{5.03} \\
\bottomrule
\end{tabular}
\end{adjustbox}
\end{table}

\begin{figure}[t]
  \centering
  \setlength{\tabcolsep}{0.5pt}\renewcommand{\arraystretch}{0.6}
  \begin{tabular}{@{}ccccc@{}}
    \includegraphics[width=0.193\linewidth]{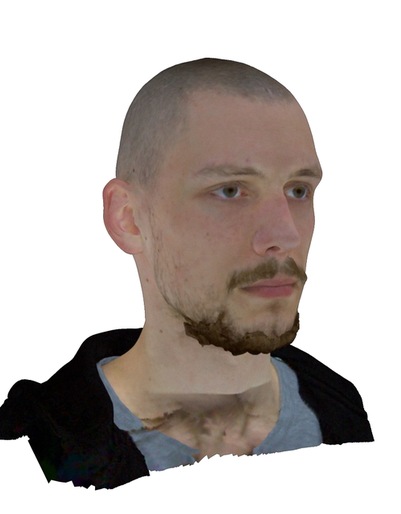} & \includegraphics[width=0.193\linewidth]{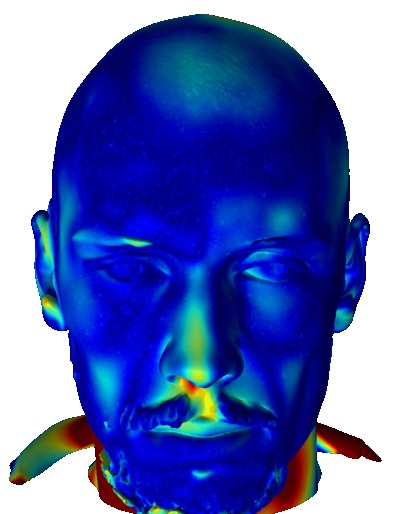} & \includegraphics[width=0.193\linewidth]{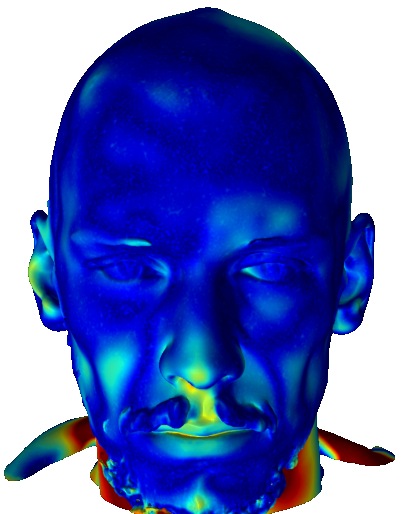} & \includegraphics[width=0.193\linewidth]{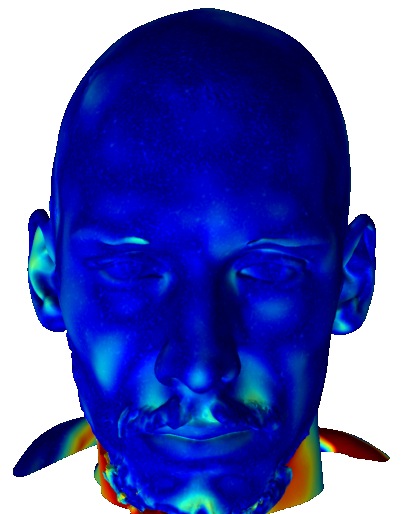} & \includegraphics[width=0.193\linewidth]{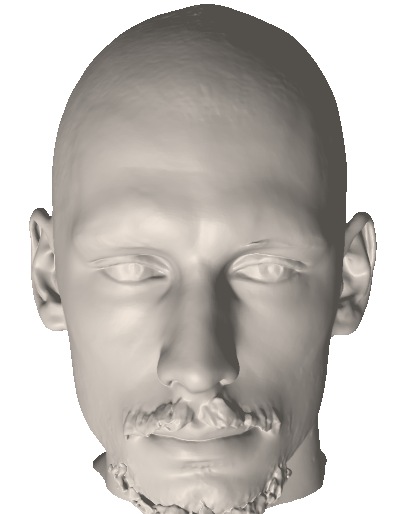} \\
    \scriptsize Input & \scriptsize 1 image & \scriptsize 2 images & \scriptsize 4 images & \scriptsize GT scan
  \end{tabular}\\[3pt]
  {\scriptsize 0\,mm}\,\raisebox{-0.2ex}{\includegraphics[width=0.5\linewidth,height=0.7em]{fig/qual/colorbar_jet.png}}\,{\scriptsize ${\geq}8$\,mm}
  \caption{\textbf{More images.} Head error of one NPHM frame, generated from its input image alone and together with one and three other images of the same person.}
  \label{fig:supp_nsweep_strip}
\end{figure}

\paragraph{Number of images.}
\Cref{tab:supp_nsweep} lists the head metrics of \cref{fig:nsweep}, and \cref{fig:supp_nsweep_strip} shows the head error of one target frame generated from one, two and four images.
Every target is generated with the first $n{-}1$ frames of a fixed random order of the other frames of its identity, and the fitted correspondences are recomputed for every $n$.
With more images, the head becomes more accurate.
\begin{table}[t]
  \centering
  \caption{\textbf{Number of images.} Head metrics of the target frames of \cref{fig:nsweep}, with distances in mm and F-score in \%.}
  \label{tab:supp_nsweep}
  \footnotesize
  \begin{tabular}{@{}lcccc@{}}
\toprule
Images & Chamfer$\downarrow$ & Compl.$\downarrow$ & EMD$\downarrow$ & F-score$\uparrow$ \\
\midrule
1 & 3.87 & 3.95 & 13.3 & 44.3 \\
2 & 3.68 & 3.71 & 12.5 & 46.9 \\
4 & 3.25 & 3.26 & 12.0 & 50.5 \\
8 & 3.23 & 3.23 & 12.1 & 50.9 \\
\bottomrule
\end{tabular}

\end{table}

\FloatBarrier
\section{Ablations}
\label{sec:supp_ablation}
All ablations are evaluated on the 80 NPHM frames with the protocol of \cref{tab:main}, the same test-time fits and the same seeds as \ours.
Every variant changes one setting of its reference, and both stages are retrained with otherwise identical settings unless the variant needs no training.

\paragraph{Design choices.}
\Cref{tab:supp_abl_design} varies the joint stages, the keys, the training correspondences and the backbone training.
Generating both stages jointly performs best.
Most of the gain comes from joint generation in the structure stage, since S1 determines the occupied voxels and thus the hair volume and back of the head.
Joint generation in the shape stage further improves the result.
Without the depth features or the null key, the overall accuracy is similar, but the lips, the region whose shape changes most between expressions, become less accurate.
Training on the fitted pool alone performs slightly worse than mixed training.
Freezing the backbone also captures slightly less facial detail than fine-tuning it jointly with the adapters.
\begin{table}[t]
  \centering
  \caption{\textbf{Design choices on NPHM.} Distances in mm; groups of rows are separated by rules.}
  \label{tab:supp_abl_design}
  \footnotesize\setlength{\tabcolsep}{2.5pt}
  \begin{adjustbox}{max width=\linewidth}\begin{tabular}{@{}lccccc@{}}
\toprule
Variant & Chamfer$\downarrow$ & Compl.$\downarrow$ & EMD$\downarrow$ & Face$\downarrow$ & Lips$\downarrow$ \\
\midrule
Joint structure stage only & 3.31 & 3.25 & 12.2 & 1.430 & \underline{1.603} \\
Joint shape stage only & 3.82 & 3.76 & 13.1 & 1.465 & 1.644 \\
\midrule
No depth features & \textbf{3.17} & \underline{3.21} & \textbf{11.7} & \textbf{1.386} & 1.735 \\
No null key & 3.33 & 3.23 & \textbf{11.7} & 1.418 & 1.756 \\
\midrule
Fitted-only training & 3.32 & 3.27 & \textbf{11.7} & 1.440 & 1.698 \\
\midrule
Frozen backbone & 3.30 & \textbf{3.20} & 12.0 & 1.427 & 1.645 \\
\midrule
Ours & \underline{3.27} & 3.23 & \underline{11.8} & \underline{1.393} & \textbf{1.565} \\
\bottomrule
\end{tabular}
\end{adjustbox}
\end{table}

\paragraph{Training compute.}
Our adapters are trained jointly with the backbone for additional steps.
For comparison, we fine-tune \pixft without adapters using the same number of steps, trainer, data, group sizes and images per step, then generate each frame independently (\cref{tab:supp_abl_compute}).
The additional fine-tuning does not improve \pixft and even slightly reduces its accuracy.
This indicates that our method's gains come from joint generation rather than additional training.
\begin{table}[t]
  \centering
  \caption{\textbf{Matched training compute on NPHM.} The metrics of \cref{tab:main}, with distances in mm and F-score in \%.}
  \label{tab:supp_abl_compute}
  \footnotesize\setlength{\tabcolsep}{2.5pt}
  \begin{adjustbox}{max width=\linewidth}\begin{tabular}{@{}lccccc@{}}
\toprule
Variant & Chamfer$\downarrow$ & Compl.$\downarrow$ & EMD$\downarrow$ & F-score$\uparrow$ & Cons.$\downarrow$ \\
\midrule
\pixft (one-by-one) & \underline{3.86} & \underline{3.77} & \underline{13.3} & \underline{45.6} & 6.73 \\
\pixft + matched fine-tuning & 4.04 & 3.91 & \underline{13.3} & 44.8 & \underline{6.67} \\
\midrule
Ours & \textbf{3.27} & \textbf{3.23} & \textbf{11.8} & \textbf{50.9} & \textbf{5.03} \\
\bottomrule
\end{tabular}
\end{adjustbox}
\end{table}

\FloatBarrier
\section{Runtime and Memory}
\label{sec:supp_runtime}
We measure the time of every step for the eight images of each of \nRtsIds NPHM identities on one NVIDIA A100, running \pixft and our method one after the other in the same job.
\Cref{tab:supp_runtime} gives the mean time per image of every step and the peak GPU memory.
Our method takes \nRtsOurs\,s per image on average and \pixft \nRtsPix\,s, about \nRtsFactor times as long.
\begin{table}[t]
  \centering
  \caption{\textbf{Runtime and memory for eight images.} Mean time per image in seconds over \nRtsIds NPHM identities, and peak GPU memory.}
  \label{tab:supp_runtime}
  \footnotesize
  \begin{tabular}{@{}lcc@{}}
\toprule
Step & \pixft & Ours \\
\midrule
Single-image generation & 27 & 30 \\
Joint generation & -- & 31 \\
Pixel3DMM & -- & 21 \\
FLAME fit & -- & 15 \\
MonoNPHM group fit & -- & 62 \\
Mesh decimation & -- & 14 \\
Clean-up & 87 & 81 \\
Other & -- & 6 \\
\midrule
Total time per image (s) & 114 & 260 \\
Peak GPU memory (GB) & 20 & 46 \\
\bottomrule
\end{tabular}

\end{table}

\FloatBarrier
\section{Additional Qualitative Results}
\label{sec:supp_qual}
\begin{figure*}[p]
  \input{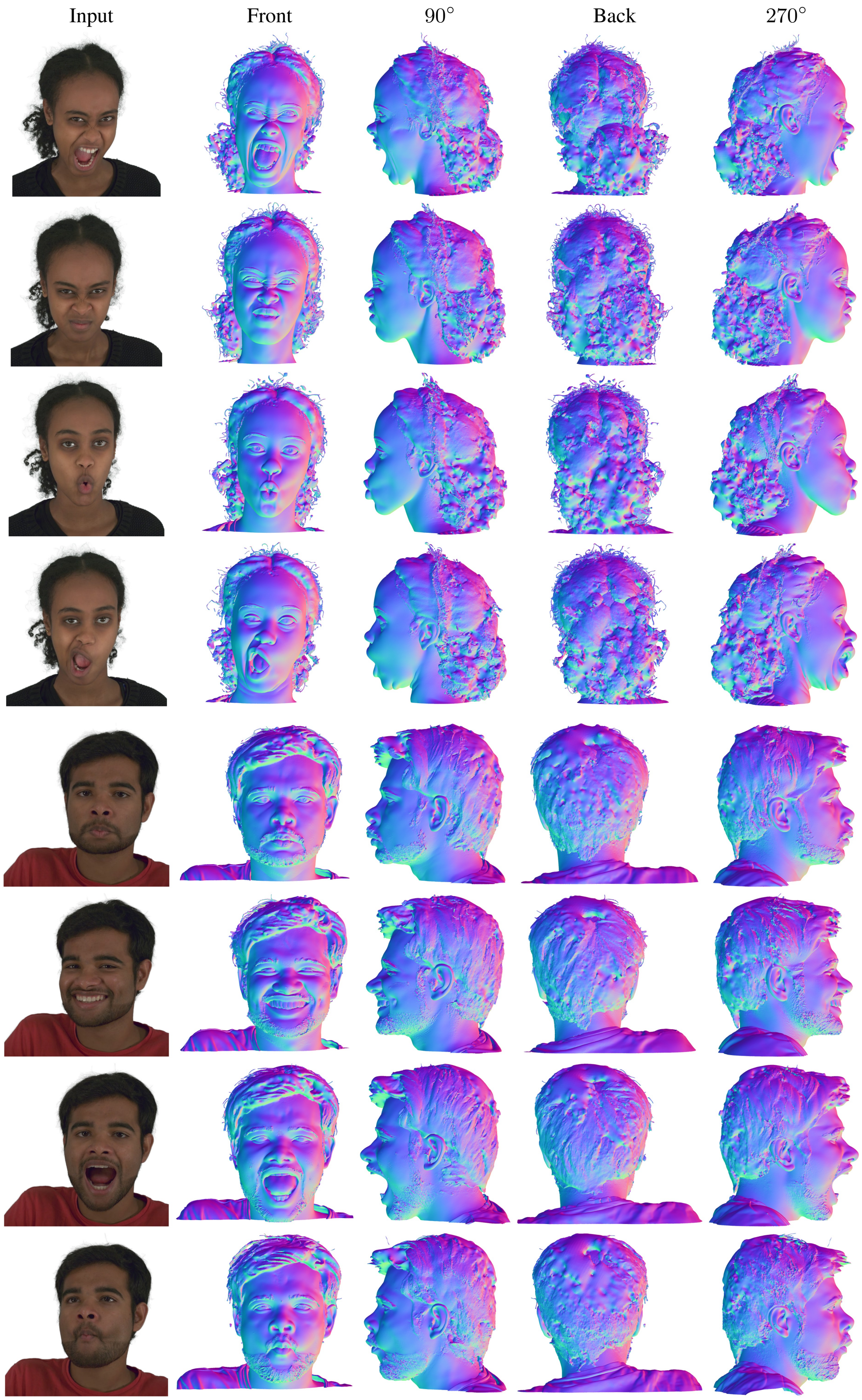}
  \caption{\textbf{More results on NeRSemble.} Two identities with four input images each.}
  \label{fig:supp_qual_ns}
\end{figure*}

\begin{figure*}[p]
  \input{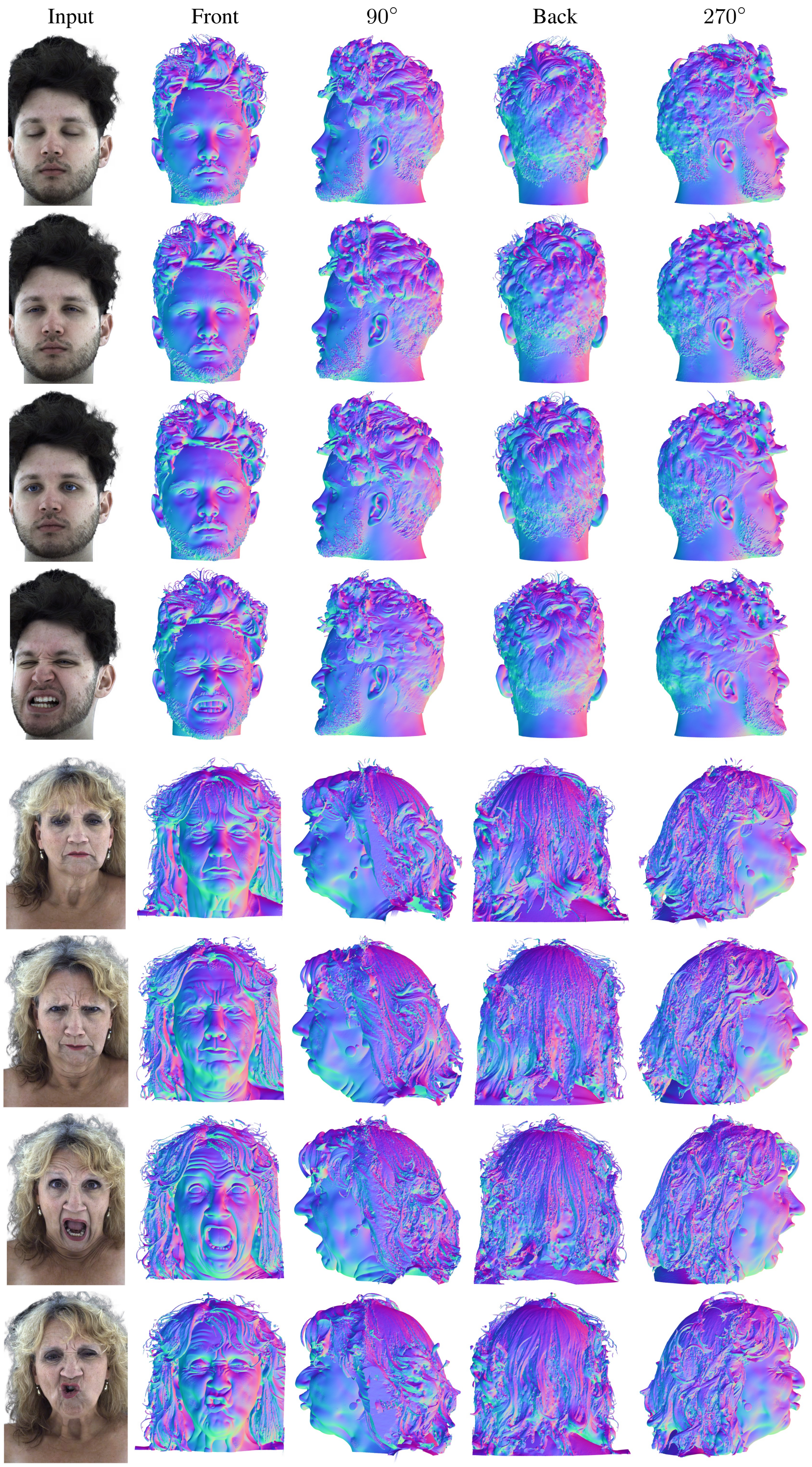}
  \caption{\textbf{More results on AVA-256.} Two identities with four input images each.}
  \label{fig:supp_qual_ava}
\end{figure*}

\begin{figure*}[p]
  \input{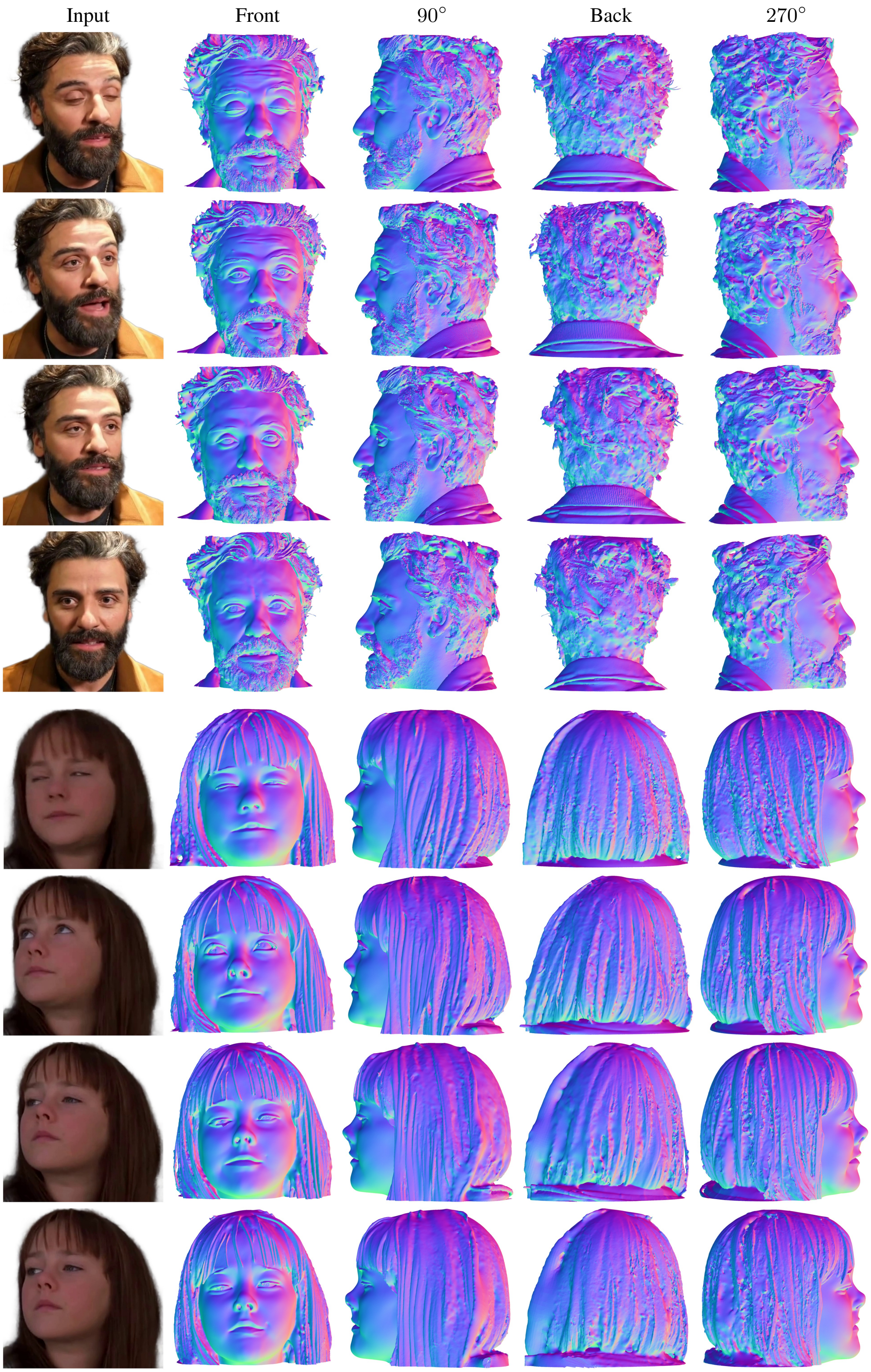}
  \caption{\textbf{More results on CelebV-HQ.} Two identities with four frames each from one video.}
  \label{fig:supp_qual_cv}
\end{figure*}

\begin{figure*}[t]
  \input{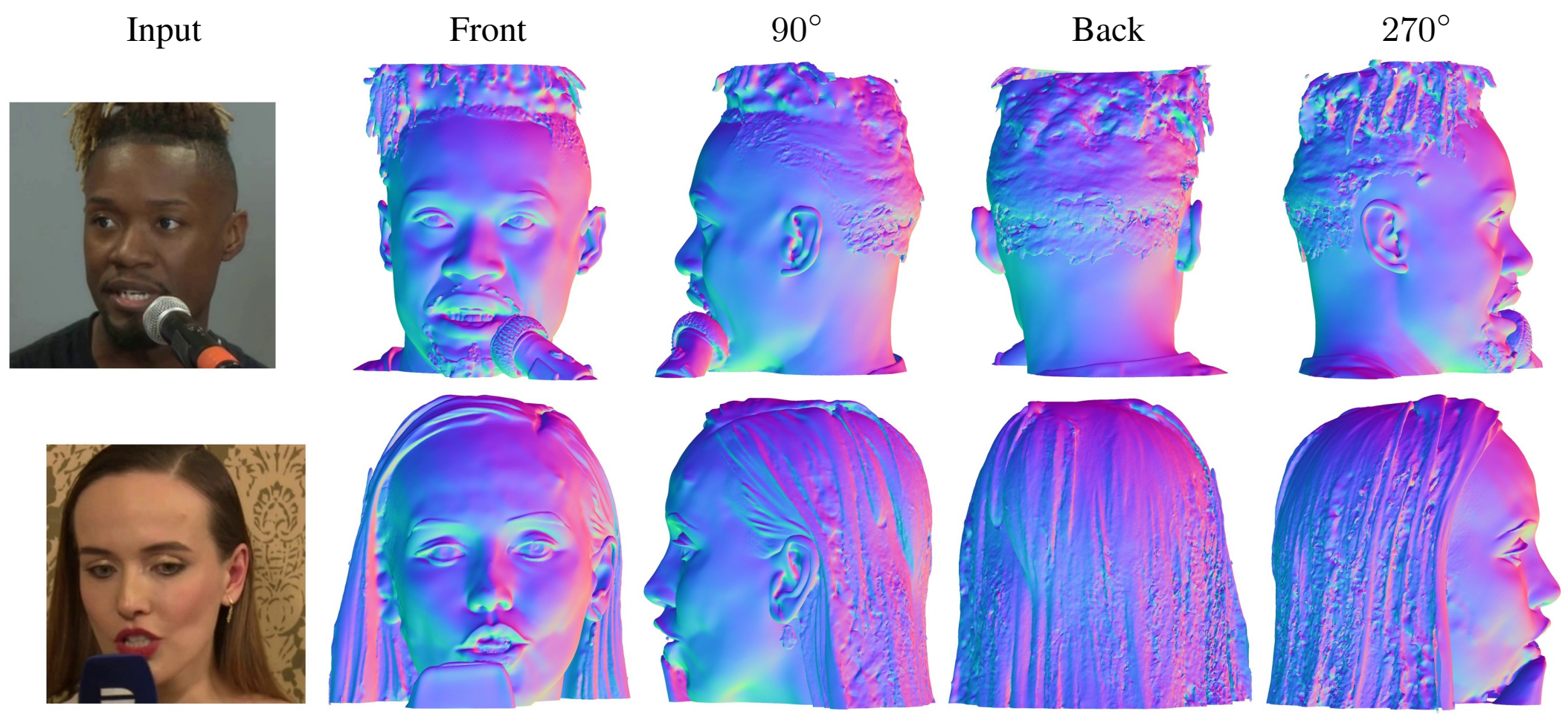}
  \caption{\textbf{Failure cases.} A microphone held in front of the mouth or at the chin is joined with the head mesh, and the microphone and the face overlap in one surface.}
  \label{fig:supp_failure}
\end{figure*}

\Cref{fig:supp_qual_ns}, \cref{fig:supp_qual_ava}, \cref{fig:supp_qual_vf} and \cref{fig:supp_qual_cv} show further results of our method on NeRSemble~\cite{kirschstein2023nersemble}, AVA-256~\cite{martinez2024ava256}, VFHQ~\cite{xie2022vfhq} and CelebV-HQ~\cite{zhu2022celebvhq}, two identities with four images each.
Our method produces high-quality, detailed head geometry from studio captures and in-the-wild video frames alike.
Every mesh is shown in the head frame from the front, from both sides and from the back, in surface-normal colours.
\Cref{fig:supp_failure} shows two failure cases with a hand-held microphone in front of the mouth or at the chin.
Our method merges the microphone and head into one surface rather than separating them.
The microphone also overlaps with the mouth and chin.
The rest of the head, including the back, stays plausible and consistent with the input.

\begin{figure*}[p]
  \input{fig/supp_qual/vfhq.tex}
  \caption{\textbf{More results on VFHQ.} Two identities with four frames each from one video.}
  \label{fig:supp_qual_vf}
\end{figure*}

\end{document}